\documentclass{article} 
\usepackage{iclr2027_conference,times}

\usepackage{amsmath,amsfonts,bm}

\def\eqref#1{equation~\ref{#1}}

\def\1{\bm{1}}

\DeclareMathAlphabet{\mathsfit}{\encodingdefault}{\sfdefault}{m}{sl}
\SetMathAlphabet{\mathsfit}{bold}{\encodingdefault}{\sfdefault}{bx}{n}

\usepackage{amsmath}
\usepackage{amssymb}
\usepackage{hyperref}
\usepackage{url}
\usepackage{booktabs}
\usepackage{colortbl}
\usepackage{graphicx}
\usepackage{algorithm}
\usepackage{algpseudocode}
\usepackage{cleveref}
\usepackage[most]{tcolorbox}
\usepackage{listings}

\newtcblisting{llmprompt}[1][]{
    enhanced,
    breakable,
    colback=gray!5,
    colframe=gray!50,
    boxrule=0.5pt,
    arc=2pt,
    left=6pt,
    right=6pt,
    top=6pt,
    bottom=6pt,
    fonttitle=\bfseries,
    title={#1},
    listing only,
    listing options={
        basicstyle=\ttfamily\small,
        breaklines=true,
        breakatwhitespace=true,
        columns=fullflexible,
        keepspaces=true,
        showstringspaces=false,
        aboveskip=0pt,
        belowskip=0pt
    }
}

\definecolor{LMOPDShade}{gray}{0.94}

\title{Lexicographic Multi-Objective On-Policy Distillation}

\author{Doseok Jang \thanks{ Work done as an intern at Cohere} \\
Cohere,\\
Mila, Université de Montréal,\\
\texttt{doseok.jang@mila.quebec}\\
\And
Jon Ander Campos\\
Cohere\\
\texttt{jonander@cohere.com}\\
\And
Youran Qi\\
Cohere\\
\texttt{youran@cohere.com}
}

\iclrfinalcopy

\begin{document}

\maketitle
\pagestyle{plain}

\begin{abstract}
Reinforcement learning from verifiable rewards (RLVR) usually optimizes answer
correctness, yet useful language-model behavior also requires high-quality
reasoning and concise responses. Existing multi-reward post-training methods
typically scalarize rewards or combine specialists without explicitly
protecting a reward priority order. This is problematic when trade-offs are
asymmetric: conciseness, for example, should not improve at the cost of
correctness. We introduce \emph{Lexicographic Multi-Objective On-Policy
Distillation} (LMOPD), a multi-teacher method for integrating reward-specialized
policies under explicit priorities. For each student rollout, LMOPD selects the
specialist for the first objective whose gate detects a deficiency, then
locally projects its centered log-policy correction to remove components that
oppose higher-priority specialists. We evaluate 30B-A3B mixture-of-experts
transformer models in two- and four-expert settings on three math benchmarks,
measuring retained specialist gains.
With two experts, LMOPD's point estimates fully retain the accuracy and
reasoning-quality gains while acquiring $46.9\%$ of the conciseness gain. With
four experts, it retains $\approx90\%$ of both the accuracy gain and
reasoning-correctness gain, compared to only $\approx57\%$ by the next best evaluated baseline. Matched four-expert
ablations show that lexicographic routing outperforms random routing and that
projection further strengthens both top-priority capabilities. Across both
scales, LMOPD preserves the highest-priority capabilities more effectively than
the existing baselines we evaluate, demonstrating the value of
explicit priorities for specialist integration.
\end{abstract}

\section{Introduction}
\label{sec:intro}

Post-training has become a critical stage in building large language models (LLMs) that can solve difficult reasoning and agentic tasks. Policy-gradient methods such as Proximal Policy Optimization (PPO) \citep{schulman2017ppo} and Group Relative Policy Optimization (GRPO) \citep{shao2024deepseekmath} enable models to improve from outcome-level feedback, and reinforcement learning from verifiable rewards (RLVR) has been especially effective in domains such as mathematics and code where correctness can be automatically verified. Modern reasoning systems consequently devote substantial post-training compute to reinforcement learning on verifiable tasks \citep{yang2025qwen3}.

Outcome correctness, however, is only one dimension of response quality. A
useful mathematical assistant should also reason correctly, verify its work,
avoid unnecessary dead ends, and communicate concisely. These objectives can
conflict and have reward signals with different scales and densities. A
standard solution is scalarization; GDPO, for example, normalizes reward
dimensions separately before combining them \citep{liu2026gdpo}. A fixed
scalarization still treats objectives as compensatory: enough improvement on a
lower-priority axis can offset degradation of a prerequisite. Many applications
instead require a non-compensatory hierarchy in which presentation quality
cannot redeem an incorrect solution.

An alternative is to specialize separate models and combine them after
training. Rewarded Soups \citep{rame2023rewardedsoups}, for example,
interpolates parameters of models optimized for different rewards. On-policy
distillation (OPD) instead transfers dense teacher feedback on student-generated
trajectories \citep{agarwal2024onpolicy}. Multi-teacher OPD (MOPD) routes
rollouts to domain-specialized teachers \citep{ma2026mopd}; there, the prompt
domain determines which teacher rollouts are routed to. We study when several objectives apply to the
\emph{same} prompt, where the missing signal is instead which requirement the
individual sampled solution fails.

We propose \textbf{Lexicographic Multi-Objective On-Policy Distillation
(LMOPD)} for priority-ordered specialist integration. LMOPD uses each rollout's
reward profile to make the routing decision. We assume an ordered set of
objectives $r_1 \succ r_2 \succ \cdots \succ r_K $
and a collection of reward-specialized experts. For each on-policy student rollout, LMOPD walks through the rewards in priority order. A discrete objective routes to its expert when its maximum acceptable reward has not been achieved; a continuous objective routes when the rollout has negative group-relative advantage on that objective. The first objective that triggers determines the teacher. If every objective passes, the rollout is routed to the student itself (effectively contributing no gradient) to avoid unnecessarily perturbing already correct behavior. Higher-priority objectives therefore have the first opportunity to correct a trajectory, while lower-priority experts only act after higher-priority requirements are satisfied.

\begin{figure}[t]
\centering
\begin{minipage}[b]{0.49\textwidth}
    \centering
    \includegraphics[
        width=\linewidth,
        trim=0 10 0 15,
        clip
    ]{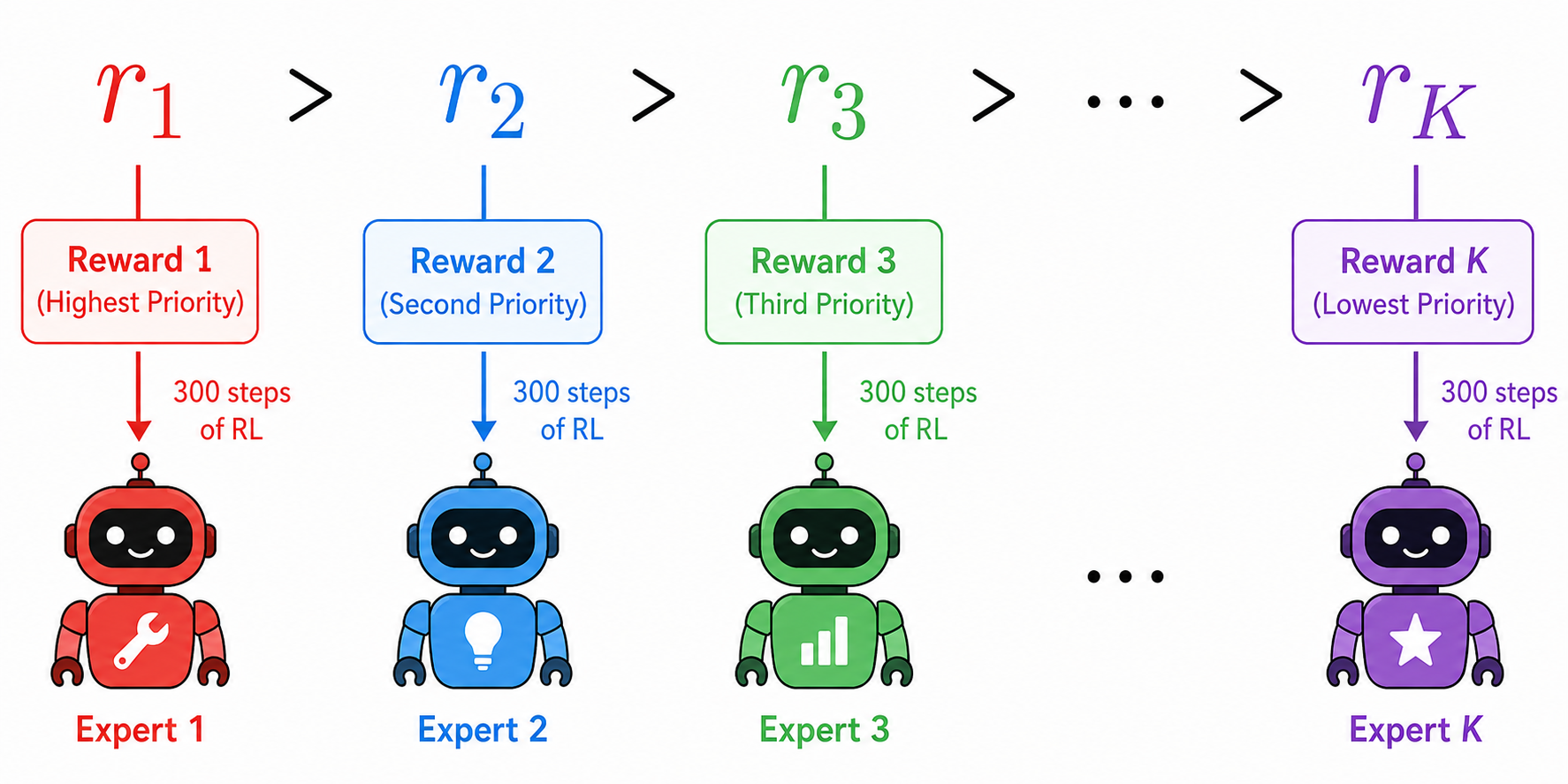}\\[-0.7ex]
    {\scriptsize\textbf{(a)} Reward-specialized experts}
\end{minipage}
\hfill
\begin{minipage}[b]{0.49\textwidth}
    \centering
    \includegraphics[
        width=\linewidth,
        trim=0 35 0 45,
        clip
    ]{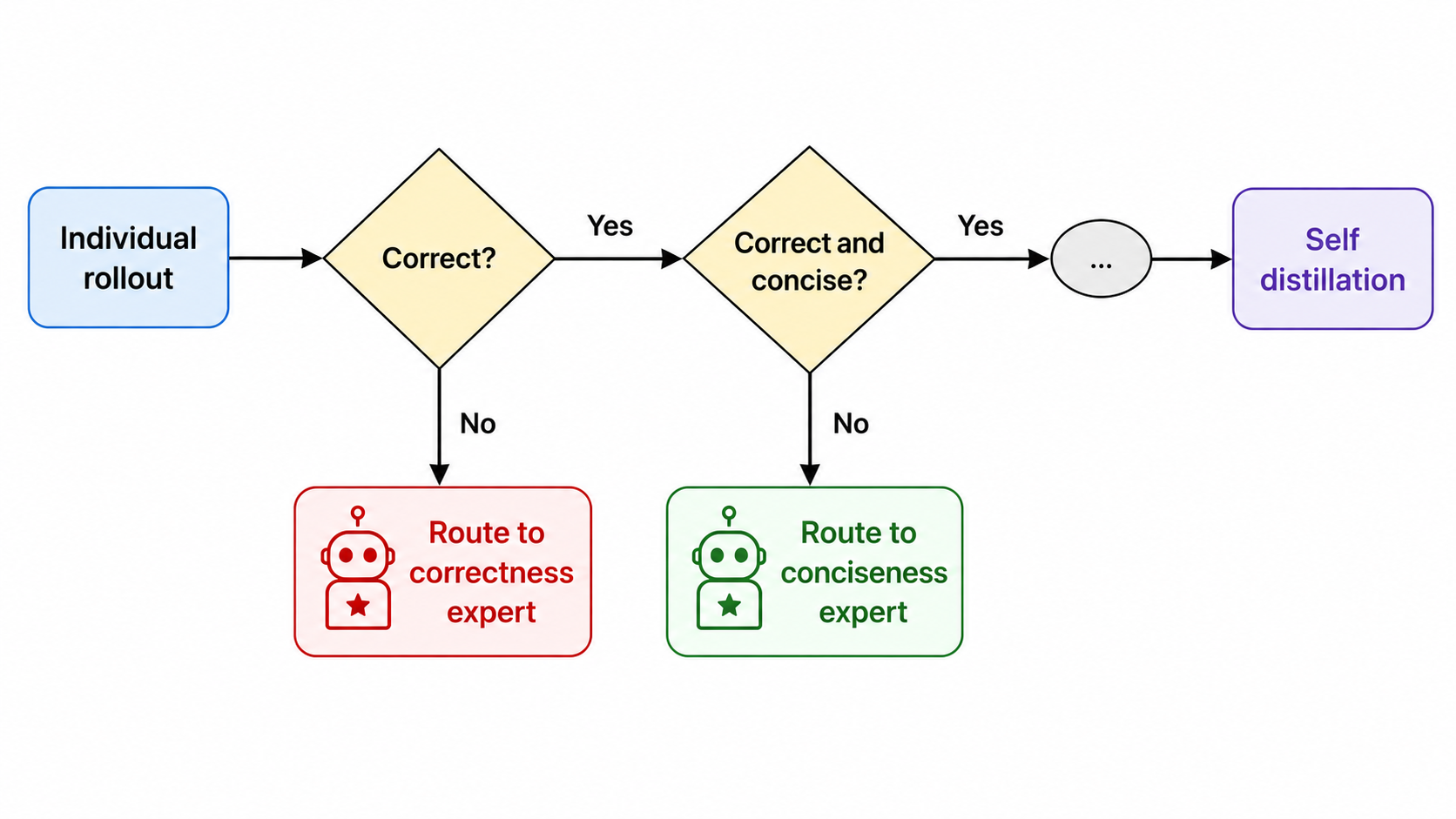}\\[-0.7ex]
    {\scriptsize\textbf{(b)} Lexicographic rollout routing}
\end{minipage}

\vspace{0.4ex}
\includegraphics[
    width=0.94\textwidth,
    trim=0 0 0 0,
    clip
]{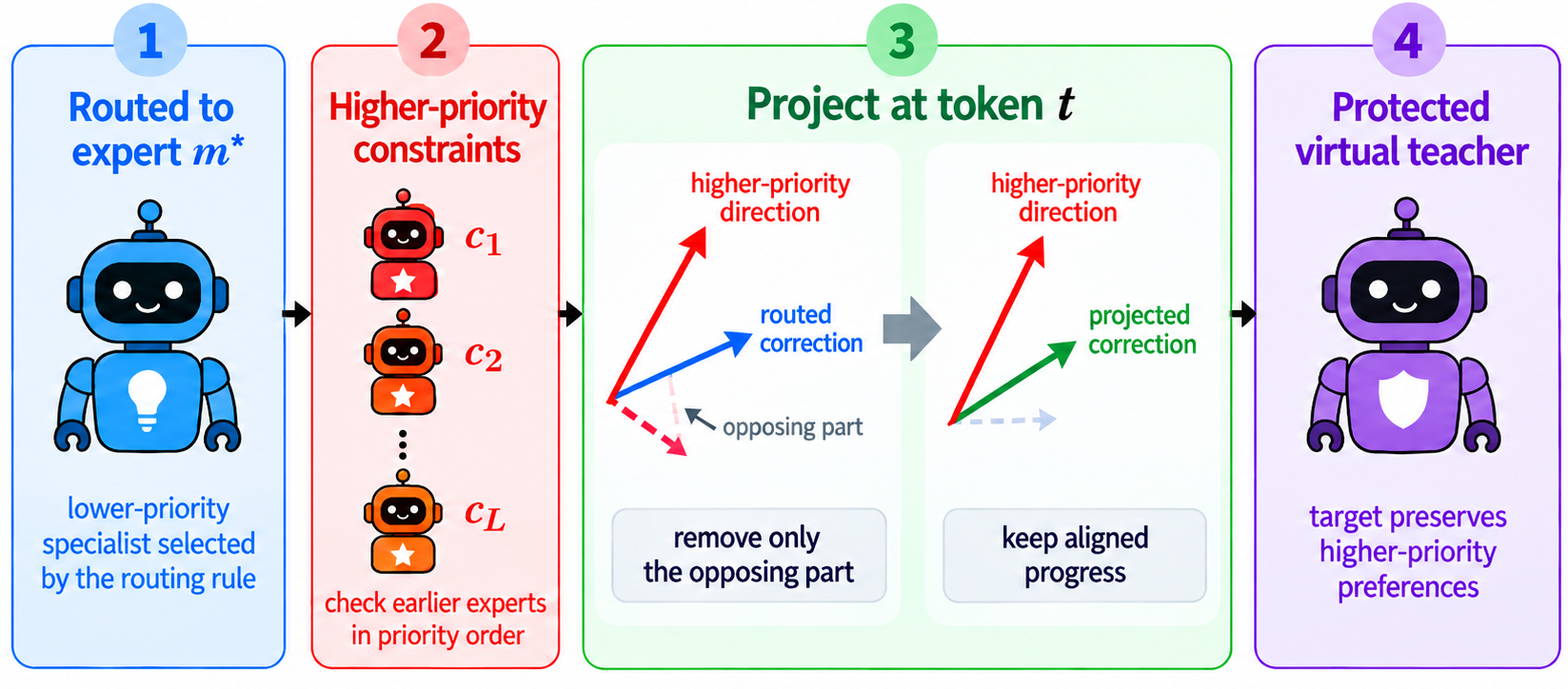}\\[-0.7ex]
{\scriptsize\textbf{(c)} Priority-aware function-space projection}
\caption{\textbf{LMOPD overview.}
\textbf{(a)} We train reward-specialized experts and assign each reward to an
expert; one expert may serve multiple rewards.
\textbf{(b)} Each student rollout is checked in that order and routed to the
first objective whose gate detects a deficiency; a rollout that passes every
gate is self-distilled.
\textbf{(c)} At each token, LMOPD removes only components of the routed
specialist's correction that oppose earlier specialists, producing a projected
virtual-teacher target.}
\label{fig:lmopd-overview}
\end{figure}

Routing alone does not prevent a selected lower-priority expert from proposing
a token-level correction that opposes a higher-priority expert. We therefore
represent each expert's correction as a centered log-policy delta and
sequentially remove components with negative inner product against the
higher-priority expert deltas. This projection is performed in function space
rather than parameter-gradient space and is implemented through streamed
per-token Gram matrices, avoiding one additional student backward pass per
expert.

This paper makes three contributions:
\begin{itemize}
    \item We formulate \emph{lexicographic corrective routing} for multi-teacher OPD. The per-rollout rule handles absolute discrete criteria and relative continuous rewards, allows several objectives to share one expert, and self-distills when no correction is required.
    \item We introduce a one-sided function-space projection that removes components of the routed correction that oppose higher-priority experts, together with a streamed Gram-space implementation that requires no extra student backward passes.
    \item We evaluate LMOPD in two- and four-expert settings at 30B scale and isolate routing and projection through component ablations.
\end{itemize}

We evaluate whether LMOPD retains prerequisite capabilities, improves teacher assignment over random routing, and redirects integration when projection activates. LMOPD is a local supervision heuristic, not a claim of globally optimal lexicographic control.

\section{Related Work}
\label{sec:related}

\paragraph{RLVR and multi-reward reinforcement learning.}
PPO \citep{schulman2017ppo} and GRPO \citep{shao2024deepseekmath} are widely used policy-gradient methods for LLM post-training. RLVR couples these updates with automatically verifiable outcome rewards, particularly for mathematical and coding tasks. Multi-reward extensions introduce additional reward dimensions for properties such as style, format, length, or process quality. GDPO \citep{liu2026gdpo} identifies a failure mode in naively applying group normalization after summing heterogeneous rewards and instead normalizes each reward dimension independently before aggregation. This preserves reward-specific variation, but the final policy update still depends on a scalarized combination of objectives. Recent work also uses conditional reward structures. SpatialThinker explicitly uses lexicographically gated rewards in multimodal RL \citep{batra2025spatialthinker}, while SCOPE-RL gates process-quality rewards on final-answer correctness in mathematical reasoning \citep{liu2026scoperl}. LMOPD shares the intuition that lower-priority feedback should not override task success, but uses this hierarchy to select a \emph{teacher distribution} rather than to construct a scalar reward.

\paragraph{Combining reward-specialized policies.}
Rewarded Soups \citep{rame2023rewardedsoups} trains policies on different
rewards from a shared initialization, linearly interpolates their weights, and uses the resulting merged model for all inputs and trajectories.

\paragraph{On-policy and multi-teacher distillation.}
Knowledge distillation trains a student to match a teacher distribution;
reverse KL provides a mode-seeking signal for autoregressive LLMs
\citep{gu2024minillm}. GKD uses student-generated sequences
\citep{agarwal2024onpolicy} instead, providing feedback that is both dense and on-policy. MOPD extends GKD by dispatching rollouts to
domain-specialized teachers to mix capabilities across domains and multiple teachers\citep{ma2026mopd}. MOPD has emerged as a key component of frontier LLM post-training recipes, including MiMo-V2-Flash \cite{xiao2026mimo}, DeepSeek-V4 \cite{xu2026deepseek}, GLM-5 \cite{zeng2026glm}, and Nemotron-Cascade 2 \cite{yang2026nemotron}. SMOPD, a concurrent work, first trains
reward specialists, then merges a token-level mixture of all teachers with a
balanced multi-reward policy-gradient anchor \citep{wang2026smopd}.

At the mechanism level, LMOPD is related to PCGrad, which removes pairwise
conflicting parameter-gradient components in multi-task learning
\citep{yu2020gradient}. LMOPD instead filters one routed teacher correction
only against earlier priorities, using an unweighted token-level function-space
surrogate rather than parameter gradients.

\paragraph{Lexicographic optimization.}
Lexicographic multi-objective reinforcement learning optimizes objectives according to an ordered preference relation rather than treating all trade-offs as interchangeable. Recent work studies this setting theoretically and algorithmically, including regret guarantees for linear MDPs \citep{xue2025lexicographic} and thresholded lexicographic formulations \citep{tercan2024thresholded}. LMOPD does not attempt to provide the same policy-level optimality guarantees. Instead, it borrows the priority structure to select a corrective teacher signal and locally project that signal during distillation.

Across these lines of work, LMOPD differs in three coupled choices: unlike
domain-routed MOPD, it routes individual rollouts by reward deficiency; unlike
Rewarded Soups and SMOPD, it does not globally or simultaneously merge all
specialists; and unlike GDPO, it does not collapse priorities into a scalarized
advantage.

\section{Method}
\label{sec:method}

LMOPD has two conceptually separable components. First, \emph{lexicographic
rollout routing} uses the reward profile of each student-generated completion
to select the specialist associated with its highest-priority deficiency.
Second, \emph{priority-aware projection} modifies the selected
specialist's token-level correction whenever it opposes a higher-priority
specialist. This decomposition yields a routing-only method and a
routing-plus-projection method with the same rollout assignments and
distillation loss. Figure~\ref{fig:lmopd-overview} summarizes the complete
pipeline.

\subsection{Lexicographic Multi-Objective Setting}

Let $x \sim \mathcal{D}$ denote a prompt and let a student policy $\pi_\theta$ generate a completion $y=(y_1,\ldots,y_T) \sim \pi_\theta(\cdot\mid x)$. We evaluate a rollout with $K$ reward functions
\begin{equation}
    \mathbf{r}(x,y) = \left(r_1(x,y),\ldots,r_K(x,y)\right),
\end{equation}
with priority ordering $r_1 \succ r_2 \succ \cdots \succ r_K$. For a policy $\pi$, define performance on objective $i$ as
\begin{equation}
    J_i(\pi) = \mathbb{E}_{x\sim\mathcal{D},\,y\sim\pi(\cdot\mid x)}[r_i(x,y)].
\end{equation}
The ordering encodes an asymmetric preference: improvement on a lower-priority
objective is not exchangeable for degradation of a higher-priority objective.

We are additionally given $M$ frozen specialists $\{\pi_{\phi_1},\ldots,\pi_{\phi_M}\}$. Let $e(i)\in\{1,\ldots,M\}$ map reward $i$ to the expert used to correct that objective. The mapping need not be one-to-one: if one specialist is strong on several correlated rewards, multiple $i$ may share the same $e(i)$.

\subsection{On-Policy Distillation}

At the start of an update, we freeze an anchor
$\bar\pi_\theta\leftarrow\pi_\theta$, sample student rollouts, and minimize a
token-averaged reverse KL toward a fixed target distribution together with a hard-label cross entropy term \citep{agarwal2024onpolicy, hinton2015distilling}. The target is the routed specialist, the anchor
itself, or the projected virtual teacher introduced below. Unlike domain-based
MOPD \citep{ma2026mopd}, LMOPD chooses this target from the rollout's reward
profile. Appendix~\ref{sec:opd-details} gives the complete objective and
notation.

\subsection{Stage I: Lexicographic Rollout Routing}
\label{sec:routing}

We sample $G$ rollouts for each prompt from the frozen anchor, as in
group-based RL methods:
\begin{equation}
    y_{b,g}\sim\bar\pi_\theta(\cdot\mid x_b),
    \qquad b\in\{1,\ldots,B\},\quad g\in\{1,\ldots,G\}.
\end{equation}
Writing $r_{b,g,i}=r_i(x_b,y_{b,g})$, we compute the within-prompt moments
and group-relative advantage:
\begin{equation}
\mu_{b,i}=\frac{1}{G}\sum_g r_{b,g,i},
\qquad
\sigma_{b,i}=\sqrt{\frac{1}{G}\sum_g(r_{b,g,i}-\mu_{b,i})^2},
\qquad
A_{b,g,i}=\frac{r_{b,g,i}-\mu_{b,i}}{\sigma_{b,i}+\epsilon}.
\label{eq:group-advantage}
\end{equation}

A single satisfaction rule is inappropriate for every reward type. For
discrete rewards with a known maximum acceptable value $r_i^{\max}$, we use
an absolute gate; for continuous rewards, we use relative underperformance:
\begin{equation}
g_i(x_b,y_{b,g})=\mathbb{I}[r_{b,g,i}<r_i^{\max}]
\ \text{(discrete)},
\qquad
g_i(x_b,y_{b,g})=\mathbb{I}[A_{b,g,i}<0]
\ \text{(continuous)}.
\label{eq:reward-gates}
\end{equation}
The absolute gate is important for sparse binary rewards: if every rollout in
a group is incorrect, all group-relative advantages are zero, yet every
rollout still requires correctness supervision.

LMOPD selects the first objective whose gate fires:
\begin{equation}
k^*(x_b,y_{b,g})
=
\min\left\{i\in\{1,\ldots,K\}:g_i(x_b,y_{b,g})=1\right\}.
\label{eq:first-failure}
\end{equation}
If the set is nonempty, the routed expert is
$m^*=e(k^*)$. If every gate passes, no expert is selected and the rollout is
self-distilled. Let $\beta_{m^*}\in(0,1]$ be the routed-teacher coefficient
used by the existing OPD logit pool; $\beta_{m^*}=1$ recovers a direct
specialist target. We use $\beta_{m^*}=0.9$ for all experiments. Writing
$p_{0,b,g,t}=\bar\pi_\theta(\cdot\mid s_{b,g,t})$ and
$p_{m,b,g,t}=\pi_{\phi_m}(\cdot\mid s_{b,g,t})$, the routing-only target is
\begin{equation}
q^{\mathrm{route}}_{b,g,t}
=
\begin{cases}
\operatorname{softmax}\!\left(
(1-\beta_{m^*})\log p_{0,b,g,t}
+
\beta_{m^*}\log p_{m^*,b,g,t}
\right),&
\text{if at least one gate fires},\\
p_{0,b,g,t},&
\text{otherwise}.
\end{cases}
\label{eq:teacher-selection}
\end{equation}
Substituting Eq.~\ref{eq:teacher-selection} into
Eq.~\ref{eq:reverse-kl} gives the routing-only LMOPD ablation.
The ordering therefore gives each objective a ``first right of refusal'' over
the rollout: incorrect responses receive capability supervision before
lower-priority properties such as conciseness are considered. For projection,
routing also records the ordered set $\mathcal{C}(k^*)$ of distinct
higher-priority experts, excluding the routed expert itself.
Appendix~\ref{sec:routing-details} gives the exact construction, edge cases,
and pseudocode.

\subsection{Stage II: Priority-Aware Function-Space Projection}
\label{sec:projection}

\paragraph{Why routing alone is insufficient.}
Routing controls which expert supervises a rollout, but it does not control
the direction of that expert's update. A lower-priority expert can therefore
assign a correction that locally opposes a higher-priority capability. LMOPD
tests compatibility with an inner product between token-level
corrections. The test concerns only the teacher-dependent
reverse-KL component and excludes the hard-label cross-entropy term. This surrogate avoids an additional student backward pass per expert compared to computing the inner product on the actual gradients;
more details can be found in Appendix~\ref{sec:projection-details}.

\paragraph{Centered log-policy corrections.}
For a token prefix $s_t$, write the frozen student anchor as
$p_{0,t}=\bar\pi_\theta(\cdot\mid s_t)$ and expert $m$ as
$p_{m,t}=\pi_{\phi_m}(\cdot\mid s_t)$. For every vocabulary item $a$, define
\begin{equation}
\Delta_{m,t}(a)
=\log p_{m,t}(a)-\log p_{0,t}(a),
\qquad
\widetilde{\Delta}_{m,t}(a)
=\Delta_{m,t}(a)
-\mathbb{E}_{a'\sim p_{0,t}}\!\left[\Delta_{m,t}(a')\right].
\label{eq:centered-policy-delta}
\end{equation}
$\widetilde{\Delta}_{m,t}(a)$ is just a centered version of $\Delta_{m,t}(a)$.

\paragraph{One-sided priority-aware projection.}
Suppose the rollout is routed to $m^*$ and
$\mathcal{C}(k^*)=(c_1,\ldots,c_L)$. We initialize $d_t^{(0)}
=
\beta_{m^*}\widetilde{\Delta}_{m^*,t}$,
using the same routed-teacher coefficient as
Eq.~\ref{eq:teacher-selection}. We then visit the constraint experts once in
priority order. Step $\ell$
is
\begin{equation}
u_{\ell,t}
=\left\langle d_t^{(\ell-1)},\widetilde{\Delta}_{c_\ell,t}\right\rangle,
\qquad
d_t^{(\ell)}
=d_t^{(\ell-1)}
-\frac{\min(u_{\ell,t},0)}
{\|\widetilde{\Delta}_{c_\ell,t}\|_2^2+\epsilon}
\widetilde{\Delta}_{c_\ell,t}.
\label{eq:one-sided-projection}
\end{equation}
Each step removes only an opposing component, but a single sweep may leave
conflicts among non-orthogonal constraints. This is a token-level
surrogate, not a parameter-gradient or reward-level guarantee;
Appendix~\ref{sec:projection-details} gives the exact limiting cases.
The projected virtual teacher is
\begin{equation}
q^{\mathrm{proj}}_{b,g,t}
=
\operatorname{softmax}
\left(
\log p_{0,t}+\tau d_t^{(L)}
\right),
\label{eq:projected-teacher}
\end{equation}
where the projection strength is $\tau=1$ in our experiments. The complete
method substitutes Eq.~\ref{eq:projected-teacher} into
Eq.~\ref{eq:reverse-kl}, with the target and projection coefficients detached
from the student gradient.
Appendix~\ref{sec:projection-details} gives the complete algorithm, limiting
cases, and one-sweep feasibility analysis. Appendix~\ref{sec:gram-projection}
describes the streamed Gram-space implementation and numerical safeguards.

\section{Experiments}
\label{sec:experiments}

\subsection{Training}
\paragraph{Datasets.}
We use a proprietary mathematical-reasoning corpus for post-training.

\paragraph{Reward functions.}
\label{sec:rewards}
We use binary answer correctness, reasoning quality judged by a 31B
Gemma 4 judge \citep{gemmateam2026gemma4technicalreport} as an average over four axes (reasoning correctness, verification, clarity, and efficiency), and a bounded length-based score $S_{\mathrm{conc}}$. All rewards are in the interval $[0,1]$. Every reported Conc metric is computed
from raw $S_{\mathrm{conc}}$. The conciseness specialist is trained with a
correctness-gated version where it only receives $S_{\mathrm{conc}}$ on correct rollouts and $0$ otherwise, whereas GDPO directly optimizes raw
$S_{\mathrm{conc}}$. Appendix~\ref{app:reward-details} gives the complete definitions.

The orders encode prerequisites rather than interchangeable utilities.
Correctness comes first because presentation cannot redeem a wrong answer;
reasoning correctness follows as a core capability, then verification,
efficiency, and conciseness refine otherwise correct solutions. We therefore read results from highest to lowest priority and treat their macro-average as a descriptive summary.

\paragraph{Expert models.}
We train reward-specialized experts from the shared base and select teachers by target-reward validation performance. This produces the two integration settings below; See Appendix~\ref{app:specialist-details} for further details on specialist training and expert selection.

\textit{Two-expert setting (three rewards).}
\label{sec:two-expert}

The first setting asks whether a student can acquire a lower-priority behavior
from one specialist while retaining the core mathematical capability of
another. We order the objectives as
$R_C \succ R_{\mathrm{RQ}} \succ R_{\mathrm{conc}}$ and use two teachers:
\begin{enumerate}
    \item a capability expert, selected from the reasoning-specialized runs and used for both the correctness and aggregate reasoning-quality gates; and
    \item a conciseness expert, used for the conciseness gate.
\end{enumerate}
Because both higher-priority gates share one capability teacher, this setting isolates adding conciseness while retaining core mathematical capability. As we use one teacher for both $R_C$ and $R_{\mathrm{RQ}}$, rollouts that route to either one of those gates would get the same loss.

\textit{Four-expert setting (five rewards).}
\label{sec:four-expert}

The second setting tests a longer priority chain with supervision distributed
across more specialists. We use
$R_C \succ R_{\mathrm{RC}} \succ R_{\mathrm{RV}} \succ R_{\mathrm{RE}}
\succ R_{\mathrm{conc}}$. The four teachers are:
\begin{enumerate}
    \item a reasoning-correctness expert, shared by the correctness and reasoning-correctness gates;
    \item a reasoning-verification expert;
    \item a reasoning-efficiency expert; and
    \item a conciseness expert.
\end{enumerate}
This setting tests integration under a longer chain in which distinct
specialist corrections compete and later teachers act only after earlier gates pass. It therefore tests both multi-expert capability integration and the scaling behavior of the lexicographic routing rule.

\paragraph{Student model.}
The LMOPD student and all RL-trained specialist and baseline policies start
from the same roughly 30B-parameter mixture-of-experts checkpoint,
with approximately 3B active parameters per token. \ificlrfinal
The checkpoint is an internal predecessor of Cohere's North Mini Code
\citep{cohere_north_code_mini}.
\else
The model architecture is closely related to a subsequently released
open-weight model; additional details will be provided in the de-anonymized
version.
\fi

\paragraph{Baselines.}
\label{sec:baselines}

We compare LMOPD with baselines that isolate different components of the method.

\textit{Rewarded Soup.}
Rewarded Soups \citep{rame2023rewardedsoups} linearly interpolates the parameters of reward-specialized policies trained from the same initialization. We evaluate both a uniform interpolation and a correctness-emphasized interpolation selected without using held-out test sets. This tests whether capability integration can be achieved by a static weight-space merge rather than by on-policy teacher supervision.

\textit{GDPO.}
GDPO \citep{liu2026gdpo} is our primary multi-objective RL baseline. It
optimizes raw correctness, aggregate reasoning quality, and raw
$S_{\mathrm{conc}}$, normalizing each reward dimension before combining the
advantages into one policy-gradient direction. GDPO uses objective weights,
rather than hand-designed gates, to control the trade-off. We evaluate equal
weights and correctness-emphasized $5{:}1$ and $25{:}1$ weightings; the main
tables show equal weights and the best-tuned $25{:}1$ run, while
\Cref{tab:raw_scores} retains the intermediate configuration.

\textit{Correctness-Gated RLVR.}
As a lexicographic reward-design control, Correctness-Gated RLVR makes
correctness a prerequisite by zeroing the aggregate reasoning-quality and
conciseness rewards on incorrect rollouts, then scalarizes the rewards that
remain. Appendix~\ref{app:baseline-details} gives both baseline objectives.
Because $R_{\mathrm{RQ}}$ is the mean of the same four reasoning-quality axes,
Correctness-Gated RLVR and each GDPO weighting receive the same three aggregate
rewards in the two- and four-expert settings. We therefore reuse the same
step-300 policy checkpoints in both tables; the four-expert table only
decomposes aggregate reasoning quality into component axes for evaluation.

\textit{Individual specialists.}
We report the frozen reward-specialized experts themselves. These models are not unified multi-objective solutions; instead, they establish the per-objective headroom available from specialization and define the reference point for retained gain.

\textit{Component ablations.}
The routing-only ablation is exactly the same as LMOPD but skips the projection step. This isolates the contribution of priority-aware projection. We also present a random-routing control which is exactly the same as the routing-only variant but matches experts to rollouts at random. This tests whether improvements come merely from exposure to multiple teachers or from matching teacher identity to the rollout's reward deficiencies and priority order.

\paragraph{Training settings.}
Two- and four-expert LMOPD variants train for 40 and 80 steps, respectively,
and use their final checkpoints. All RL-trained specialist and baseline
policies use their final step-300 checkpoint.
See Appendix~\ref{app:experimental-details} for further details.

\subsection{Evaluation}

\paragraph{Datasets and protocol.}
We generate 50 completions per problem on AIME 2024, AIME 2025, and HMMT
February 2025 and report the average and standard error of the mean (SEM) None of these benchmarks is used for training or checkpoint
selection.
See Appendix~\ref{app:experimental-details} for further details.

\paragraph{Metrics and uncertainty.}
\label{sec:metrics}
We aggregate each reported objective across the three benchmarks. We report
pass@1, aggregate reasoning quality (plus the three relevant component axes), and raw $S_{\mathrm{conc}}$ independent of correctness. Raw
objective values are means with standard errors of the mean (SEM), while
retained-gain SEs use the delta method.

For objective $i$, let $J_i(\pi)$ denote the model's mean performance, let $\pi_0$ be the shared base policy, and let $\pi_i^\star$ be the best available specialist for that objective. We report the \emph{retained gain}
\begin{equation}
G_i(\pi)
=
\frac{
J_i(\pi)-J_i(\pi_0)
}{
J_i(\pi_i^\star)-J_i(\pi_0)
}
\times 100\%.
\label{eq:retained-gain}
\end{equation}
Retained gain measures how much of the improvement created by specialization survives after capabilities are integrated into one model. A value of $0\%$ means that the unified method does not improve over the base on that objective, while $100\%$ means that it closes the entire base-to-specialist gap; values above $100\%$ indicate that the integrated policy surpasses the best specialist, and negative values indicate degradation relative to the base. This normalization is useful because reward axes can differ substantially in both scale and available headroom. We report the corresponding raw objective values in \Cref{tab:raw_scores} and treat the macro-average retained gain as a descriptive summary rather than a substitute for inspecting high-priority objectives individually.

\section{Results}
\label{sec:results}

\subsection{Two-Expert Retained Gain}
\Cref{tab:retained_gain_2expert_composite} reports retained gain aggregated
over AIME 2024, AIME 2025, and HMMT February 2025, with 50 sampled
completions per problem. LMOPD preserves the two high-priority capabilities:
its point estimates retain $(102.9 \pm 15.5)\%$ of the accuracy gain
(raw: $0.711\!\to\!0.745$) and $(103.3 \pm 5.7)\%$ of the aggregate
reasoning-quality gain (raw: $0.719\!\to\!0.772$), while retaining
$(46.9 \pm 2.2)\%$ of the conciseness gain. Because $100\%$ denotes the
best-specialist endpoint, both high-priority estimates are consistent within
their SEs with specialist-level performance.

LMOPD has the highest performance among unified methods
($84.4\%$), $17.3$ points above the strongest non-LMOPD baseline, the
correctness-emphasized Rewarded Soup ($67.1\%$). That soup retains substantial
gains in aggregate reasoning quality and conciseness but only $72.7\%$ of the
available accuracy gain. Equal-weight GDPO reduces mean completion length from $16.3$k
tokens for the base to $6.2$k while lowering raw accuracy from $0.711$ to
$0.559$. At $25{:}1$, GDPO avoids this collapse: mean length rises to
$13.3$k tokens, and raw accuracy and aggregate reasoning quality recover to
$0.700$ and $0.717$, respectively, while retaining $90.5\%$ of the
conciseness gain. The equal-weight run is therefore consistent with
length-dominant reward hacking, while stronger correctness weighting
substantially mitigates it.
\begin{table}[t]
\centering
\caption{Two-expert retained gain. LMOPD has the highest average among unified
methods while retaining gains in accuracy and aggregate reasoning quality.}
\label{tab:retained_gain_2expert_composite}
\scriptsize
\resizebox{\textwidth}{!}{%
\begin{tabular}{lcccc}
\toprule
\textbf{Model} & \textbf{pass@1} & \textbf{RQ} & \textbf{Conc} & \textbf{Average} \\
\midrule
\multicolumn{5}{l}{\textit{Weight-space merging}} \\
Rewarded Soup (uniform; 1:1)
& $61.2 \pm 13.7$\% & $61.6 \pm 5.0$\% & $67.2 \pm 2.3$\% & 63.3\% \\
Rewarded Soup (correctness-emphasized; 2:1)
& $\underline{72.7 \pm 14.0}$\% & $\underline{84.3 \pm 5.3}$\% & $44.4 \pm 2.1$\% & \underline{67.1\%} \\
\addlinespace[2pt]
\multicolumn{5}{l}{\textit{Multi-reward reinforcement learning}} \\
Correctness-Gated RLVR (1,1,1)
& $70.0 \pm 13.8$\% & $40.2 \pm 5.0$\% & $49.1 \pm 2.2$\% & 53.1\% \\
GDPO (1,1,1)
& $-457.9 \pm 85.2$\% & $-154.4 \pm 15.3$\% & $\mathbf{355.8 \pm 9.8}$\% & -85.5\% \\
GDPO (25,1,1)
& $-31.5 \pm 20.1$\% & $-2.4 \pm 4.7$\% & $\underline{90.5 \pm 3.7}$\% & 18.9\% \\
\addlinespace[2pt]
\multicolumn{5}{l}{\textit{Ours}} \\
\rowcolor{LMOPDShade}
\textbf{LMOPD}
& $\mathbf{102.9 \pm 15.5}$\% & $\mathbf{103.3 \pm 5.7}$\% & $46.9 \pm 2.2$\% & \textbf{84.4\%} \\
\midrule
\multicolumn{5}{l}{\textit{Reference policies (not unified)}} \\
Base
& 0\% & 0\% & 0\% & 0\% \\
Reasoning Quality + Correctness Expert
& 100\% & 100\% & $39.2 \pm 2.2$\% & 79.7\% \\
Conciseness Expert
& $0.8 \pm 16.1$\% & $-20.4 \pm 6.5$\% & 100\% & 26.8\% \\
\bottomrule
\end{tabular}%
}
\vspace{2pt}
\begin{minipage}{\textwidth}
\footnotesize
Bold and underlined entries are the best and second-best unified methods in
each column; reference policies are excluded. LMOPD denotes routing plus
priority-aware projection and uses the fixed step-40 checkpoint. Best specialist
per axis: pass@1 = Reasoning Quality + Correctness Expert (0.7440), RQ =
Reasoning Quality + Correctness Expert (0.7698), and Conc = Conciseness Expert
(0.8131). The shared base defines $0\%$.
\end{minipage}
\end{table}

For qualitative analysis of LMOPD's completions, we present a case study of an individual eval problem in Appendix~\ref{app:qualitative}.


\subsection{Four-Expert Retained Gain}
The four-expert setting tests whether high-priority capabilities remain
protected as the priority chain grows to five objectives.
\Cref{tab:retained_gain_4expert_composite} shows that LMOPD retains
$(90.6 \pm 15.0)\%$ of the accuracy gain (raw: $0.711\!\to\!0.748$) and
$(88.6 \pm 7.0)\%$ of the reasoning-correctness gain
(raw: $0.755\!\to\!0.805$). Every other unified method reaches at most
$57.1\%$ and $63.5\%$, respectively. LMOPD also has the highest macro-average
point estimate ($39.6\%$, versus $34.4\%$ for Correctness-Gated RLVR).

Because the objectives are lexicographically ordered, the columns should be
read in priority order rather than as equally consequential targets. The
central scaling result is therefore LMOPD's large margin on accuracy and
reasoning correctness, followed by a positive composite-conciseness gain
($28.0 \pm 1.1$\%). Reasoning verification and efficiency are lower-priority
objectives: their retained gains of $-6.8\%$ and $-2.4\%$ do not outweigh
preservation of the first two objectives under the stated preference order.
This profile is consistent with the intended behavior of LMOPD---preserving
high-priority capabilities even when all specialist directions cannot be
improved simultaneously. \textbf{LMOPD outperforms other multi-objective optimization baselines on important objectives.}
\begin{table}[t]
\centering
\caption{Four-expert retained gain, with objectives ordered by priority.
LMOPD leads unified methods on composite pass@1, RQ-corr, and average.}
\label{tab:retained_gain_4expert_composite}
\scriptsize
\resizebox{\textwidth}{!}{%
\begin{tabular}{lcccccc}
\toprule
\textbf{Model} & \textbf{Composite pass@1} & \textbf{RQ-corr} & \textbf{RQ-verif} & \textbf{RQ-eff} & \textbf{Composite Conc} & \textbf{Average} \\
\midrule
\multicolumn{7}{l}{\textit{Weight-space merging}} \\
Rewarded Soup (uniform; 1:1:1:1)
& $31.4 \pm 14.3$\% & $51.6 \pm 7.5$\% & $-2.3 \pm 2.5$\% & $28.7 \pm 2.1$\% & $40.2 \pm 1.4$\% & 29.9\% \\
Rewarded Soup (correctness-emphasized; 2:1:1:1)
& $30.5 \pm 11.6$\% & $\underline{63.5 \pm 10.4}$\% & $\mathbf{4.2 \pm 2.9}$\% & $19.2 \pm 2.0$\% & $28.5 \pm 0.8$\% & 29.2\% \\
\addlinespace[2pt]
\multicolumn{7}{l}{\textit{Multi-reward reinforcement learning}} \\
Correctness-Gated RLVR (1,1,1)
& $\underline{57.1 \pm 9.5}$\% & $55.7 \pm 7.7$\% & $\underline{3.2 \pm 2.4}$\% & $30.8 \pm 4.4$\% & $25.2 \pm 1.8$\% & \underline{34.4\%} \\
GDPO (1,1,1)
& $-373.4 \pm 55.7$\% & $-265.0 \pm 35.7$\% & $-161.5 \pm 6.1$\% & $\mathbf{95.1 \pm 2.0}$\% & $\mathbf{182.1 \pm 3.0}$\% & -104.6\% \\
GDPO (25,1,1)
& $-25.7 \pm 14.0$\% & $5.6 \pm 9.1$\% & $-35.6 \pm 3.6$\% & $\underline{36.0 \pm 2.4}$\% & $\underline{46.3 \pm 1.9}$\% & 5.3\% \\
\addlinespace[2pt]
\multicolumn{7}{l}{\textit{Ours}} \\
\rowcolor{LMOPDShade}
\textbf{LMOPD}
& $\mathbf{90.6 \pm 15.0}$\% & $\mathbf{88.6 \pm 7.0}$\% & $-6.8 \pm 2.7$\% & $-2.4 \pm 2.8$\% & $28.0 \pm 1.1$\% & \textbf{39.6\%} \\
\midrule
\multicolumn{7}{l}{\textit{Reference policies (not unified)}} \\
Base
& 0\% & 0\% & 0\% & 0\% & 0\% & 0\% \\
RQ-Correctness Expert
& 100\% & 100\% & $7.7 \pm 3.2$\% & $-8.9 \pm 3.2$\% & $13.0 \pm 0.9$\% & 42.4\% \\
RQ-Verification Expert
& $79.7 \pm 12.5$\% & $124.4 \pm 9.5$\% & 100\% & $-58.5 \pm 4.4$\% & $-44.3 \pm 2.1$\% & 40.3\% \\
RQ-Efficiency Expert
& $-202.8 \pm 33.7$\% & $-118.5 \pm 20.7$\% & $-106.5 \pm 4.9$\% & 100\% & 100\% & -45.6\% \\
Conciseness Expert
& $0.7 \pm 12.8$\% & $25.3 \pm 9.9$\% & $-47.6 \pm 3.3$\% & $21.7 \pm 2.1$\% & $51.2 \pm 1.2$\% & 10.2\% \\
\bottomrule
\end{tabular}%
}
\vspace{2pt}
\begin{minipage}{\textwidth}
\footnotesize
Bold and underlined entries are the best and second-best unified methods in
each column; reference policies are excluded. LMOPD denotes routing plus
priority-aware projection. Best specialist
per axis: pass@1 = RQ-Correctness Expert (0.7515), RQ-corr = RQ-Correctness Expert
(0.8117),
RQ-verif = RQ-Verification Expert (0.7902), RQ-eff = RQ-Efficiency Expert
(0.7827), and Conc = RQ-Efficiency Expert (0.8402).
\end{minipage}
\end{table}


\subsection{Component Ablations}
\Cref{tab:lmopd_component_ablations} separates teacher assignment from
priority-aware projection. Lexicographic routing's average retained-gain exceeds random routing by $28.7$ points with two experts and $23.0$
points with four. In the four-expert setting, random routing improves later
objectives but falls below the base on accuracy and reasoning correctness;
teacher exposure alone is therefore insufficient.

Projection exposure differs sharply by setting. In the two-expert run, only
$0.6\%$ of rows carry a higher-priority constraint, corresponding to
$0.003$--$0.005$ constraints per token and a projected-to-original norm ratio
of $0.9989$. Projection is therefore effectively inert there, so the
performance shifts from routing only to full LMOPD are
modest relative to their uncertainty. In the four-expert run, projection is materially engaged:
it processes $0.26$ constraints per token, adds constraint forwards for
$13.7\%$ of rows, and yields a norm ratio of $0.91$. At the matched step-80
checkpoint, projection shifts the retained-gain point estimates for accuracy
and reasoning correctness from $54.0\%$ to $90.6\%$ and from $69.5\%$ to $88.6\%$,
respectively, while reducing later-axis gains.

\begin{table}[t]
\centering
\caption{Routing and projection ablations. Lexicographic routing has higher
performance than random routing; matched four-expert projection shifts
retention toward the top priorities.}
\label{tab:lmopd_component_ablations}
\scriptsize
\textbf{(a) Two experts}\par\vspace{2pt}
\resizebox{\textwidth}{!}{%
\begin{tabular}{lcccc}
\toprule
\textbf{Variant} & \textbf{pass@1} & \textbf{RQ} & \textbf{Conc} & \textbf{Average} \\
\midrule
\rowcolor{LMOPDShade}
\textbf{LMOPD}
& $102.9 \pm 15.5$\% & $103.3 \pm 5.7$\% & $46.9 \pm 2.2$\% & 84.4\% \\
LMOPD (routing only)
& $95.3 \pm 12.5$\% & $97.3 \pm 4.8$\% & $43.5 \pm 1.8$\% & 78.7\% \\
LMOPD (random routing)
& $30.4 \pm 14.0$\% & $46.8 \pm 4.2$\% & $72.8 \pm 2.0$\% & 50.0\% \\
\bottomrule
\end{tabular}%
}

\vspace{2pt}
\textbf{(b) Four experts}\par\vspace{2pt}
\resizebox{\textwidth}{!}{%
\begin{tabular}{lcccccc}
\toprule
\textbf{Variant} & \textbf{Composite pass@1} & \textbf{RQ-corr} & \textbf{RQ-verif} & \textbf{RQ-eff} & \textbf{Composite Conc} & \textbf{Average} \\
\midrule
\rowcolor{LMOPDShade}
\textbf{LMOPD}
& $90.6 \pm 15.0$\% & $88.6 \pm 7.0$\% & $-6.8 \pm 2.7$\% & $-2.4 \pm 2.8$\% & $28.0 \pm 1.1$\% & 39.6\% \\
LMOPD (routing only)
& $54.0 \pm 13.2$\% & $69.5 \pm 11.1$\% & $-18.4 \pm 2.9$\% & $4.3 \pm 3.4$\% & $39.7 \pm 1.0$\% & 29.8\% \\
LMOPD (random routing)
& $-53.7 \pm 14.2$\% & $-2.0 \pm 13.6$\% & $-39.6 \pm 3.6$\% & $53.8 \pm 2.7$\% & $75.6 \pm 1.6$\% & 6.8\% \\
\bottomrule
\end{tabular}%
}
\vspace{2pt}
\begin{minipage}{\textwidth}
\end{minipage}
\end{table}

\section{Limitations}
\label{sec:discussion} LMOPD serves as a local heuristic rather than global lexicographic optimality: its function-space test approximates parameter-gradient compatibility, and one sweep may leave residual conflicts.

Strict routing can starve later experts
(Appendix~\ref{app:routing-starvation}), and outcomes ultimately depend
on reward and teacher quality. We evaluate only proprietary-model mathematical
reasoning; open-weight and cross-domain validation remain future work. Because
30B-scale post-training is costly, multi-seed replication is limited to
two-expert LMOPD: three 40 step runs (see Appendix~\ref{sec:multiseed-replication}). The main-table run is
the median seed by average retained gain ($84.4\%$; the other seeds obtain
$81.2\%$ and $95.1\%$);
\Cref{tab:multiseed_2expert,fig:multiseed_loss} give the complete results.
Except for this replication study, reported SEs quantify evaluation sampling,
not training-run variation.

\section{Conclusion}
\label{sec:conclusion}

We introduced LMOPD, which routes each rollout to the specialist for its
highest-priority unmet objective and projects that specialist's token-level
correction against earlier experts. With two experts, LMOPD retains
specialist-level accuracy and aggregate reasoning quality while acquiring conciseness;
with four, it retains $90.6\%$ of the accuracy gain and $88.6\%$ of the
gain in reasoning correctness, with higher point estimates than multi-objective baselines on both
top priorities. Ablations show that deficiency-aware routing is essential and
that, when materially engaged with four experts, projection shifts point
estimates toward these priorities. Together, these results position
rollout-conditioned distillation as a practical path to integrating specialists without collapsing prioritized capabilities into a single global trade-off.

\subsection*{AI use statement}

In this work, we used generative AI tools for coding assistance in implementing methods, designing and providing feedback on methods and experiments, refining hypotheses, and supporting qualitative data analysis. 
We have not used generative AI tools for interpreting results, cleaning and reformatting datasets,
and generating synthetic datasets, developing theoretical models or frameworks, formulating mathematical claims, providing critical ingredients for proving mathematical claims, writing of proofs, translation are not applicable to this work.
Additionally, we used generative AI tools for modifying scientific figures, editing the research paper, analyzing existing literature. We have reviewed all AI-assisted work. LLM-generated code, experiment feedback, and paper modifications were verified by the main author for correctness. We take responsibility for the final content of this work,
including text, claims or artifacts produced with the aid of generative AI.

%
%

\subsection*{Reproducibility statement}
The LMOPD routing rule is given in Eqs.~\ref{eq:group-advantage}--\ref{eq:teacher-selection} and Algorithm~\ref{alg:lmopd}; the priority-aware target is specified in Eqs.~\ref{eq:centered-policy-delta}--\ref{eq:projected-teacher}, Algorithm~\ref{alg:priority-projection}, and the Gram-space implementation in Appendix~\ref{sec:gram-projection}; and the distillation objective is specified in Eqs.~\ref{eq:reverse-kl}--\ref{eq:opd-objective}. The experimental setup, baselines, and retained-gain metric are summarized in Section~\ref{sec:experiments}, with reward definitions, training schedules, batch construction, and evaluation details in Appendix~\ref{app:experimental-details}. The base checkpoint and training corpus used in our experiments are proprietary, so those exact assets cannot be released. The method itself requires only a student policy, white-box access to reward-specialized teacher distributions, and the stated reward functions, and can therefore be reproduced with open-weight models by training specialists from a common initialization. The paper provides implementation-level pseudocode and configuration details sufficient to reproduce the algorithm independently of the proprietary checkpoint.

%

\bibliography{iclr2027_conference}

@article{schulman2017ppo,
  title   = {Proximal Policy Optimization Algorithms},
  author  = {Schulman, John and Wolski, Filip and Dhariwal, Prafulla and Radford, Alec and Klimov, Oleg},
  journal = {arXiv preprint arXiv:1707.06347},
  year    = {2017},
  url     = {https://arxiv.org/abs/1707.06347}
}

@article{shao2024deepseekmath,
  title   = {{DeepSeekMath}: Pushing the Limits of Mathematical Reasoning in Open Language Models},
  author  = {Shao, Zhihong and Wang, Peiyi and Zhu, Qihao and Xu, Runxin and Song, Junxiao and Bi, Xiao and Zhang, Haowei and Zhang, Mingchuan and Li, Y. K. and Wu, Y. and Guo, Daya},
  journal = {arXiv preprint arXiv:2402.03300},
  year    = {2024},
  url     = {https://arxiv.org/abs/2402.03300}
}

@article{yang2025qwen3,
  title   = {{Qwen3} Technical Report},
  author  = {Yang, An and Li, Anfeng and Yang, Baosong and Zhang, Beichen and Hui, Binyuan and Zheng, Bo and Yu, Bowen and Gao, Chang and Huang, Chengen and Lv, Chenxu and Zheng, Chujie and Liu, Dayiheng and Zhou, Fan and Huang, Fei and Hu, Feng and Ge, Hao and Wei, Haoran and Lin, Huan and Tang, Jialong and Yang, Jian and Tu, Jianhong and Zhang, Jianwei and Yang, Jianxin and Yang, Jiaxi and Zhou, Jing and Zhou, Jingren and Lin, Junyang and Dang, Kai and Bao, Keqin and Yang, Kexin and Yu, Le and Deng, Lianghao and Li, Mei and Xue, Mingfeng and Li, Mingze and Zhang, Pei and Wang, Peng and Zhu, Qin and Men, Rui and Gao, Ruize and Liu, Shixuan and Luo, Shuang and Li, Tianhao and Tang, Tianyi and Yin, Wenbiao and Ren, Xingzhang and Wang, Xinyu and Zhang, Xinyu and Ren, Xuancheng and Fan, Yang and Su, Yang and Zhang, Yichang and Zhang, Yinger and Wan, Yu and Liu, Yuqiong and Wang, Zekun and Cui, Zeyu and Zhang, Zhenru and Zhou, Zhipeng and Qiu, Zihan},
  journal = {arXiv preprint arXiv:2505.09388},
  year    = {2025},
  url     = {https://arxiv.org/abs/2505.09388}
}

@inproceedings{agarwal2024onpolicy,
  title     = {On-Policy Distillation of Language Models: Learning from Self-Generated Mistakes},
  author    = {Agarwal, Rishabh and Vieillard, Nino and Zhou, Yongchao and Stanczyk, Piotr and Garea, Sabela Ramos and Geist, Matthieu and Bachem, Olivier},
  booktitle = {International Conference on Learning Representations},
  year      = {2024},
  url       = {https://proceedings.iclr.cc/paper_files/paper/2024/hash/5be69a584901a26c521c2b51e40a4c20-Abstract-Conference.html}
}

@inproceedings{gu2024minillm,
  title     = {{MiniLLM}: Knowledge Distillation of Large Language Models},
  author    = {Gu, Yuxian and Dong, Li and Wei, Furu and Huang, Minlie},
  booktitle = {International Conference on Learning Representations},
  year      = {2024},
  url       = {https://proceedings.iclr.cc/paper_files/paper/2024/hash/8ac015d409635f196f9e3e9dcfb9a94e-Abstract-Conference.html}
}

@misc{ma2026mopd,
  title         = {{MOPD}: Multi-Teacher On-Policy Distillation for Capability Integration in {LLM} Post-Training},
  author        = {Ma, Wenhan and Wei, Jianyu and Zhao, Liang and Zhang, Hailin and Xiao, Bangjun and Li, Lei and Yang, Qibin and Gao, Bofei and Wang, Yudong and Li, Rang and Dong, Jinhao and Sui, Zhifang and Luo, Fuli},
  year          = {2026},
  eprint        = {2606.30406},
  archivePrefix = {arXiv},
  primaryClass  = {cs.CL},
  url           = {https://arxiv.org/abs/2606.30406}
}

@misc{wang2026smopd,
  title         = {{SMOPD}: Multi-Reward Reinforcement Learning via Specialize-and-Merge Online Policy Distillation},
  author        = {Wang, Wen and Bao, Jiahua and Tu, Yongsiqi and Liu, Yihao and Zhou, Haotian and Ma, Haoxuan and Zhou, Mengyu and Fan, Wenkui and He, Junwei and Jiang, Xiaoxi and Jiang, Guanjun},
  year          = {2026},
  eprint        = {2608.03092},
  archivePrefix = {arXiv},
  primaryClass  = {cs.LG},
  url           = {https://arxiv.org/abs/2608.03092}
}

@misc{liu2026gdpo,
  title         = {{GDPO}: Group Reward-Decoupled Normalization Policy Optimization for Multi-Reward {RL} Optimization},
  author        = {Liu, Shih-Yang and Dong, Xin and Lu, Ximing and Diao, Shizhe and Belcak, Peter and Liu, Mingjie and Chen, Min-Hung and Yin, Hongxu and Wang, Yu-Chiang Frank and Cheng, Kwang-Ting and Choi, Yejin and Kautz, Jan and Molchanov, Pavlo},
  year          = {2026},
  eprint        = {2601.05242},
  archivePrefix = {arXiv},
  primaryClass  = {cs.CL},
  url           = {https://arxiv.org/abs/2601.05242}
}

@inproceedings{rame2023rewardedsoups,
  title     = {Rewarded Soups: Towards Pareto-Optimal Alignment by Interpolating Weights Fine-Tuned on Diverse Rewards},
  author    = {Ram{\'e}, Alexandre and Couairon, Guillaume and Shukor, Mustafa and Dancette, Corentin and Gaya, Jean-Baptiste and Soulier, Laure and Cord, Matthieu},
  booktitle = {Advances in Neural Information Processing Systems},
  volume    = {36},
  year      = {2023},
  url       = {https://proceedings.neurips.cc/paper_files/paper/2023/hash/e12a3b98b67e8395f639fde4c2b03168-Abstract-Conference.html}
}

@inproceedings{yu2020gradient,
  title     = {Gradient Surgery for Multi-Task Learning},
  author    = {Yu, Tianhe and Kumar, Saurabh and Gupta, Abhishek and Levine, Sergey and Hausman, Karol and Finn, Chelsea},
  booktitle = {Advances in Neural Information Processing Systems},
  volume    = {33},
  pages     = {5824--5836},
  year      = {2020},
  url       = {https://proceedings.neurips.cc/paper/2020/hash/3fe78a8acf5fda99de95303940a2420c-Abstract.html}
}

@inproceedings{xue2025lexicographic,
  title     = {Multi-Objective Linear Reinforcement Learning with Lexicographic Rewards},
  author    = {Xue, Bo and Bu, Dake and Cheng, Ji and Wan, Yuanyu and Zhang, Qingfu},
  booktitle = {Proceedings of the 42nd International Conference on Machine Learning},
  pages     = {70065--70085},
  year      = {2025},
  volume    = {267},
  series    = {Proceedings of Machine Learning Research},
  publisher = {PMLR},
  url       = {https://proceedings.mlr.press/v267/xue25a.html}
}

@inproceedings{tercan2024thresholded,
  title     = {Thresholded Lexicographic Ordered Multiobjective Reinforcement Learning},
  author    = {Tercan, Alperen and Prabhu, Vinayak S.},
  booktitle = {ECAI 2024},
  pages     = {3006--3014},
  year      = {2024},
  doi       = {10.3233/FAIA240841}
}

@misc{batra2025spatialthinker,
  title         = {{SpatialThinker}: Reinforcing 3D Reasoning in Multimodal {LLM}s via Spatial Rewards},
  author        = {Batra, Hunar and Tu, Haoqin and Chen, Hardy and Lin, Yuanze and Xie, Cihang and Clark, Ronald},
  year          = {2025},
  eprint        = {2511.07403},
  archivePrefix = {arXiv},
  primaryClass  = {cs.CV},
  url           = {https://arxiv.org/abs/2511.07403}
}

@misc{liu2026scoperl,
  title         = {{SCOPE-RL}: Optimizing Reasoning Paths Before and After Success},
  author        = {Liu, Xiaojian and Xu, Han and Xia, Jianqiang and Li, Zhixuan and Xu, Ke and Dai, Yiwei and Chen, Xinran and Wu, Changwo and Li, Yuchen},
  year          = {2026},
  eprint        = {2607.11506},
  archivePrefix = {arXiv},
  primaryClass  = {cs.CL},
  url           = {https://arxiv.org/abs/2607.11506}
}

@misc{gemmateam2026gemma4technicalreport,
      title={Gemma 4 Technical Report}, 
      author={Gemma Team and Sherif El Abd and Vaibhav Aggarwal and Robin Algayres and Alek Andreev and Olivier Bachem and Ian Ballantyne and Cormac Brick and Victor Cărbune and Michelle Casbon and Mayank Chaturvedi and Aditya Chawla and Victor Cotruta and Alice Coucke and Phil Culliton and Robert Dadashi and Lucas Dixon and Mohamed Elhawaty and Utku Evci and Clément Farabet and Johan Ferret and Filippo Galgani and Sertan Girgin and Jean-Bastien Grill and Maarten Grootendorst and Jiaxian Guo and Cassidy Hardin and Yanzhang He and Steven M. Hernandez and Omri Homburger and Léonard Hussenot and Juyeong Ji and Armand Joulin and Aishwarya Kamath and Parnian Kassraie and Olivier Lacombe and Preethi Lahoti and Gaël Liu and Gus Martins and Luciano Martins and Tatiana Matejovicova and Ramona Merhej and Nikola Momchev and Sneha Mondal and Ryan Mullins and Sindhu Raghuram Panyam and Shreya Pathak and Sarah Perrin and André Susano Pinto and Etienne Pot and Angéline Pouget and Alexandre Ramé and Sabela Ramos and Douglas Reid and David Rim and Morgane Rivière and Karsten Roth and Louis Rouillard and Omar Sanseviero and Pier Giuseppe Sessa and Shane Settle and Danila Sinopalnikov and Sara Smoot and Piotr Stanczyk and Andreas Steiner and Lawrence Stewart and Ilya Tolstikhin and Michael Tschannen and Anton Tsitsulin and Nino Vieillard and Renjie Wu and Pingmei Xu and Haichuan Yang and Edouard Yvinec and Biao Zhang and Li Zhang and Joe Zou and Nicolas Aagnes and Abdelrahman Abdelhamed and Jakub Adamek and Shivani Agrawal and Shubham Agrawal and Ibrahim Alabdulmohsin and Jean Baptiste Alayrac and Uri Alon and Chandramouli Amarnath and Ankesh Anand and Chrysovalantis Anastasiou and Setareh Ariafar and François-Xavier Aubet and Kyriakos Axiotis and Federico Barbero and Joelle Barral and Alexei Bendebury and Urs Bergmann and Stanley Bileschi and Kat Black and Mathieu Blondel and Sebastian Borgeaud and Arthur Bražinskas and Ryan Burnell and Robert Busa-Fekete and Mu Cai and Daniele Calandriello and Glenn Cameron and Charlotte Caucheteux and Rahma Chaabouni and Garima Chadha and Jetha Chan and Blake Jianhang Chen and Jesse Chen and Lin Chen and Xu Chen and Derek Cheng and Tzu-hsiang Chien and Nikolai Chinaev and Yi Chou and Zhaohui Chu and Benjamin Coleman and Pooja Consul and Sam Conway-Rahman and Scott Crowell and Dylan Cutler and Vivek Dani and Samira Daruki and Anil Das and Daniel Deutsch and Nishanth Dikkala and Li Ding and Qiuhan Ding and Shenil Dodhia and Konstantin Donhauser and Tulsee Doshi and Anca Dragan and Alex Druinsky and Sahil Dua and Zoltan Egyed and Danielle Eisenbud and Daniel Eppens and Cindy Fan and Bahare Fatemi and Yassir Fathullah and Vlad Feinberg and Milen Ferev and Sebastian Flennerhag and Takumi Fujimoto and João Gabriel Oliveira and Isaac Galatzer-Levy and João Gante and Simon Geisler and Soham Ghosal and Antonious M. Girgis and Tamara von Glehn and Alec Go and Alhaad Gokhale and Alex Grills and Yiming Gu and Mayank Gupta and Pramod Gupta and Guru Guruganesh and Raia Hadsell and Hamza Harkous and Jitendra Harlalka and Demis Hassabis and Anja Hauth and Joe Heyward and Arian Hosseini and Chih-Yang Hsia and I-Hung Hsu and Xiaopeng Huang and Yangsibo Huang and Kevin Hui and Adrian Hutter and Te I and Fotis Iliopoulos and Advait Jain and Ganesh Jawahar and Ziwei Ji and Qilin Jin and Melvin Johnson and Kandarp Joshi and Arun Kandoor and Wang-Cheng Kang and Koray Kavukcuoglu and Mehran Kazemi and Kathleen Kenealy and Amr Khalifa and Phoebe Kirk and Ivan Korotkov and Suraj Kothawade and Vitaly Kovalev and Neel Kovelamudi and Adam Kraft and Ravin Kumar and Vivek Kumar and Harish Kuppam and Justin Lannin and Chen-Yu Lee and Seungji Lee and Dmitry Lepikhin and Alon Levkovitch and Dongdong Li and Qiujia Li and Valentin Liévin and Ethan Lin and Ziqian Lin and Casper Liu and Tianlin Liu and Tianqi Liu and Xin Liu and Ivan Lobov and Mayank Lunayach and Min Ma and Gagan Madan and Andrii Maksai and Eric Malmi and Michal Matuszak and Daniel McDuff and Gaurav Menghani and Maciej Mikuła and Daniil Mirylenka and Karolis Misiunas and Vedant Misra and Andreea Mitran and Kareem Mohamed and Maksim Mukha and Eric Noland and James O'Donnell and Brendan O'Donoghue and Kate Olszewska and Bernett Orlando and Wanqiong Pan and Rina Panigrahy and Unnati Parekh and Nicolas Perez-Nieves and Chunjong Park and Eric Paskie and Liqian Peng and Bryce Petrini and Slav Petrov and Jonas Pfeiffer and Bilal Piot and Martyna Plomecka and Siim Poder and Octavio Ponce and Arijit Pramanik and David Racz and Anish Rajan and Michelle Ramanovich and Anand Rao and Marvin Ritter and Vitor Rodrigues and Evan Rosen and Mikołaj Rybiński and Noveen Sachdeva and Michaël E. Sander and Rohit Sathyanarayana and Sagar Savla and Samuel Schmidgall and Tal Schuster and George Scrivener and Benoit Seguin and Andrew Sellergren and Aliaksei Severyn and Izhak Shafran and Dhruv Shah and Bobak Shahriari and Yuan Shangguan and Ashish Shenoy and Pradeep Shenoy and Rakesh Shivanna and Pauline Sho and Lucas Spangher and Wojciech Stokowiec and Tim Strother and Yao Su and Yinghao Sun and Mukund Sundararajan and Andrea Tacchetti and Mor Hazan Taege and Pouya Tafti and Jean Tarbouriech and Chetan Tekur and Shantanu Thakoor and Rahul Thapa and Madeleine Traverse and Lenart Treven and Tao Tu and Chien Te Tung and Çağlar Ünlü and Petar Veličković and Malini Pooni Venkat and Sagar Gubbi Venkatesh and Vidya Venkiteswaran and Francesco Visin and Alex Vitvitskyi and Kiran Vodrahalli and Weiyi Wang and Xin Wang and Tris Warkentin and Jan Wassenberg and John Wieting and Cindy Wu and Lechao Xiao and Hao Xu and Yuhui Xu and Fuzhao Xue and Arun Yadav and Jun Yan and Antoine Yang and Lin Yang and Ming-Hsuan Yang and Ziyu Ying and Jae Hyeon Yoo and Morteza Zadimoghaddam and Sajjad Zafar and Fred Zhang and Jiageng Zhang and Jianyi Zhang and Xiaofan Zhang and Chao Zhao and David Zhou and Chen Zou},
      year={2026},
      eprint={2607.02770},
      archivePrefix={arXiv},
      primaryClass={cs.CL},
      url={https://arxiv.org/abs/2607.02770}, 
}

@article{balunovic2025matharena,
  title   = {{MathArena}: Evaluating {LLM}s on Uncontaminated Math Competitions},
  author  = {Balunovi{\'c}, Mislav and Dekoninck, Jasper and Petrov, Ivo and Jovanovi{\'c}, Nikola and Vechev, Martin},
  journal = {arXiv preprint arXiv:2505.23281},
  year    = {2025},
  url     = {https://arxiv.org/abs/2505.23281}
}

@misc{hmmt2025feb,
  author       = {{Harvard-MIT Mathematics Tournament}},
  title        = {February 2025 Tournament Archive},
  year         = {2025},
  howpublished = {HMMT Archive},
  url          = {https://www.hmmt.org/www/archive/282}
}

@misc{cohere_north_code_mini,
    title = {Introducing {North Mini Code}: Cohere's First Model For Developers},
    url = {cohere.com/blog/north-mini-code},
    author = {{Team Cohere}},
    month = {June},
    year = {2026}
}

@article{yang2026nemotron,
  title={Nemotron-cascade 2: Post-training llms with cascade rl and multi-domain on-policy distillation},
  author={Yang, Zhuolin and Liu, Zihan and Chen, Yang and Dai, Wenliang and Wang, Boxin and Lin, Sheng-Chieh and Lee, Chankyu and Chen, Yangyi and Jiang, Dongfu and He, Jiafan and others},
  journal={arXiv preprint arXiv:2603.19220},
  year={2026}
}

@article{xiao2026mimo,
  title={Mimo-v2-flash technical report},
  author={Xiao, Bangjun and Xia, Bingquan and Yang, Bo and Gao, Bofei and Shen, Bowen and Zhang, Chen and He, Chenhong and Lou, Chiheng and Luo, Fuli and Wang, Gang and others},
  journal={arXiv preprint arXiv:2601.02780},
  year={2026}
}

@article{zeng2026glm,
  title={Glm-5: from vibe coding to agentic engineering},
  author={Zeng, Aohan and Lv, Xin and Hou, Zhenyu and Du, Zhengxiao and Zheng, Qinkai and Chen, Bin and Yin, Da and Ge, Chendi and Huang, Chenghua and Xie, Chengxing and others},
  journal={arXiv preprint arXiv:2602.15763},
  year={2026}
}

@article{xu2026deepseek,
  title={Deepseek-v4: Towards highly efficient million-token context intelligence},
  author={Xu, Anyi and Lin, Bangcai and Xue, Bing and Wang, Bingxuan and Xu, Bingzheng and Wu, Bochao and Zhang, Bowei and Lin, Chaofan and Dong, Chen and Ling, Chenchen and others},
  journal={arXiv preprint arXiv:2606.19348},
  year={2026}
}

@article{hinton2015distilling,
  title={Distilling the knowledge in a neural network},
  author={Hinton, Geoffrey and Vinyals, Oriol and Dean, Jeff},
  journal={arXiv preprint arXiv:1503.02531},
  year={2015}
}
\bibliographystyle{iclr2027_conference}

\appendix
\section{Additional Method Details}
\label{app:method-details}

\subsection{On-Policy Distillation Objective}
\label{sec:opd-details}

At the start of an update, we freeze an anchor
$\bar\pi_\theta\leftarrow\pi_\theta$ and sample rollouts from this snapshot.
On-policy distillation evaluates its target on prefixes produced by the
student rather than on a fixed teacher-generated corpus
\citep{agarwal2024onpolicy}. For rollout $n$, define
\begin{equation}
    s_t^{(n)} = (x^{(n)}, y_{<t}^{(n)}), \qquad
    y^{(n)} \sim \bar\pi_\theta(\cdot\mid x^{(n)}).
\end{equation}
Let $q_t^{(n)}(\cdot\mid s_t^{(n)})$ be a fixed token-level target
distribution. Depending on the method variant, it is either the routed
specialist, the anchor itself, or the projected virtual teacher. We use a
token-averaged reverse-KL distillation loss
\begin{equation}
\mathcal{L}_{\mathrm{KD}}^{(n)}
=
\frac{1}{T_n}\sum_{t=1}^{T_n}
D_{\mathrm{KL}}\!\left(
\pi_\theta(\cdot\mid s_t^{(n)})
\,\|\,
q_t^{(n)}(\cdot\mid s_t^{(n)})
\right).
\label{eq:reverse-kl}
\end{equation}
Following our training implementation, we add a regularizing hard-label cross-entropy term on
the student-generated trajectory,
\begin{equation}
\mathcal{L}_{\mathrm{hard}}^{(n)}
=
-\frac{1}{T_n}\sum_{t=1}^{T_n}
\log \pi_\theta\!\left(y_t^{(n)}\mid s_t^{(n)}\right),
\end{equation}
and optimize
\begin{equation}
\mathcal{L}_{\mathrm{OPD}}
=
\frac{1}{N}\sum_{n=1}^{N}
\left(
\mathcal{L}_{\mathrm{KD}}^{(n)}
+
\beta_{\mathrm{hard}}\mathcal{L}_{\mathrm{hard}}^{(n)}
\right),
\qquad
\theta \leftarrow \theta - \eta\nabla_\theta \mathcal{L}_{\mathrm{OPD}}.
\label{eq:opd-objective}
\end{equation}
 We use $\beta_{\mathrm{hard}}=1-\beta_{\mathrm{m*}}=0.1$ for all experiments. Reverse KL is well suited to mode-seeking generative distillation
\citep{gu2024minillm}; the hard-label cross-entropy term regularizes the update toward the
sampled on-policy behavior.

\subsection{Routing Constraint Construction}
\label{sec:routing-details}

\paragraph{Routing intuition.}
The priority ordering gives each objective a ``first right of refusal'' over
the rollout. With correctness first, an incorrect response is always corrected
by the capability expert and cannot be pushed toward conciseness merely
because it is short. Once correctness passes, the next objective is allowed to
inspect the rollout, and so on. Continuous objectives use group-relative
advantage so that the method asks whether a rollout is unusually weak on that
dimension compared with alternative completions of the same prompt.
Self-distillation prevents trajectories that already satisfy every gate from
being arbitrarily pulled toward a specialist. The resulting update is still
dense at the token level; only the identity of the teacher is selected
lexicographically.

\paragraph{Higher-priority constraint list.}
Routing records the ordered list of distinct experts associated with
objectives above $k^*$,
\begin{equation}
\mathcal{C}(k^*)
=
\operatorname{ordered\_unique}
\left(e(1),\ldots,e(k^*-1)\right)
\setminus\{m^*\}.
\label{eq:constraint-list}
\end{equation}
The order is the objective-priority order, not checkpoint order. Removing
$m^*$ is important when one expert serves multiple objectives: an expert is
never projected against itself, while other distinct experts attached to
earlier objectives remain constraints, including experts that occur between
two objectives served by $m^*$.

\begin{algorithm}[H]
\caption{Lexicographic rollout routing}
\label{alg:lmopd}
\begin{algorithmic}[1]
\Require Rollout $(x_b,y_{b,g})$; gates $g_1,\ldots,g_K$;
reward-to-expert map $e$
\State $\mathcal{C}\leftarrow()$ \Comment{ordered list of passed experts}
\For{$i=1,\ldots,K$}
    \If{$g_i(x_b,y_{b,g})=1$}
        \State $m^*\leftarrow e(i)$
        \State Remove $m^*$ from $\mathcal{C}$ if present
        \State \Return $(m^*,\mathcal{C})$
    \EndIf
    \If{$e(i)\notin\mathcal{C}$}
        \State Append $e(i)$ to $\mathcal{C}$
    \EndIf
\EndFor
\State \Return $(\varnothing,())$ \Comment{all gates pass: self-distill}
\end{algorithmic}
\end{algorithm}

\subsection{Projection Derivation and Algorithm}
\label{sec:projection-details}

\paragraph{Parameter-gradient motivation.}
Let $z_{\theta,t}$ be the student logits at prefix $s_t$,
$p_{\theta,t}=\operatorname{softmax}(z_{\theta,t})$, and
$J_t=\partial z_{\theta,t}/\partial\theta$. At the frozen anchor, define the
categorical Fisher matrix
\begin{equation}
F_{0,t}
=
\operatorname{diag}(p_{0,t})-p_{0,t}p_{0,t}^{\top}.
\end{equation}
For a detached virtual-teacher target
$q_{m,t}=\operatorname{softmax}(\log p_{0,t}+\tau d_{m,t})$, the
teacher-dependent reverse-KL descent direction at the anchor is
\begin{equation}
g_{m,t}^{\mathrm{KD}}
:=
-\left.
\nabla_\theta
D_{\mathrm{KL}}(p_{\theta,t}\,\|\,q_{m,t})
\right|_{\theta=\bar\theta}
=
\tau J_t^\top F_{0,t}d_{m,t}.
\end{equation}
Consequently, exact token-local compatibility between two such KD components
is
\begin{equation}
\left\langle g_{m,t}^{\mathrm{KD}},g_{j,t}^{\mathrm{KD}}\right\rangle
=
\tau^2 d_{m,t}^{\top}M_t d_{j,t},
\qquad
M_t=F_{0,t}J_tJ_t^\top F_{0,t}.
\label{eq:parameter-compatibility}
\end{equation}
Centering does not remove the Fisher factor: for a
$p_{0,t}$-centered direction,
$F_{0,t}d_{m,t}=\operatorname{diag}(p_{0,t})d_{m,t}$. A full
trajectory-gradient comparison additionally contains cross-token blocks
$F_{0,t}J_tJ_u^\top F_{0,u}$. The hard-label cross-entropy term is not included here because
it is independent of teacher identity; LMOPD modifies only the
teacher-dependent KD component.

Computing the parameter directions requires an expert-specific
vector--Jacobian product. LMOPD instead tests
$d_{m,t}^{\top}d_{j,t}$, thereby replacing the complete pullback metric $M_t$
with the identity and enforcing compatibility separately at each token. It is a heuristic on centered
log-policy displacements and does not guarantee parameter-gradient or
reward-level compatibility. Because its vocabulary inner product is
unweighted, low-probability and high-probability actions have equal geometric
weight even though the exact reverse-KL direction is Fisher-weighted.

\paragraph{Limiting cases.}
When $\mathcal{C}$ is empty and $\tau=1$,
Eq.~\ref{eq:projected-teacher} is exactly the routing-only logit pool in
Eq.~\ref{eq:teacher-selection}; when $\beta_{m^*}=1$, both reduce to the routed
expert distribution. Centering does not change this equality because it adds
only an action-independent constant. Rows for which every gate passes retain
all-zero expert coefficients, remain equal to $p_{0,t}$, and never enter the
projection. Consequently, disabling projection does not change rollout
selection or the unprojected target.

\begin{algorithm}[H]
\caption{Priority-aware target for one routed rollout}
\label{alg:priority-projection}
\begin{algorithmic}[1]
\Require Frozen anchor $p_0$; routed expert $m^*$; ordered constraints
$\mathcal{C}=(c_1,\ldots,c_L)$; coefficients $\beta_{m^*},\tau$
\If{$m^*=\varnothing$}
    \State \Return $q_t\leftarrow p_{0,t}$ for every token $t$
\EndIf
\For{completion token $t=1,\ldots,T$}
    \State Compute centered deltas
    $\widetilde{\Delta}_{m^*,t}$ and
    $\{\widetilde{\Delta}_{c_\ell,t}\}_{\ell=1}^L$
    using Eq.~\ref{eq:centered-policy-delta}
    \State $d_t\leftarrow
    \beta_{m^*}\widetilde{\Delta}_{m^*,t}$
    \For{$\ell=1,\ldots,L$}
        \State $u\leftarrow
        \langle d_t,\widetilde{\Delta}_{c_\ell,t}\rangle$
        \If{$u<0$}
            \State $d_t\leftarrow d_t-
            \dfrac{u}{\|\widetilde{\Delta}_{c_\ell,t}\|_2^2+\epsilon}
            \widetilde{\Delta}_{c_\ell,t}$
        \EndIf
    \EndFor
    \State $q_t\leftarrow
    \operatorname{softmax}(\log p_{0,t}+\tau d_t)$
\EndFor
\State \Return $\{q_t\}_{t=1}^T$
\end{algorithmic}
\end{algorithm}

\paragraph{One sweep versus exact feasibility.}
With at most one active constraint, the update is the exact orthogonal
projection onto the corresponding half-space, up to $\epsilon$ and numerical
guards. With two or more non-orthogonal constraints, a later step can
reintroduce a negative inner product with an earlier expert. Repeated sweeps
converge toward the intersection of the half-spaces, but our implementation
uses one sweep by default because most routed rows have zero or one
constraint.

\paragraph{Efficient implementation.}
Because the projected direction remains in the span of the expert
corrections, we implement the same function-space sweep in coefficient space
using streamed per-token $M\times M$ Gram matrices. This avoids both
vocabulary-sized correction tensors and additional student backward passes.
Appendix~\ref{sec:gram-projection} gives the algebraic construction, streaming
procedure, and numerical safeguards.

\subsection{Routing Starvation}
\label{app:routing-starvation}

An expert at position $i$ is selected only when all higher-priority gates pass
and its own gate fails. Abstractly,
\begin{equation}
\Pr[z=e(i)]
=
\Pr\!\left[g_1=0,\ldots,g_{i-1}=0,g_i=1\right].
\end{equation}
As the number of priorities grows, later experts may receive fewer rollouts,
especially early in training. This is an inherent consequence of strict
routing and suggests extensions such as minimum routing quotas, softened
priority constraints, or adaptive gate thresholds when scaling to many
objectives.

\section{Gram-Space Projection Implementation}
\label{sec:gram-projection}

The dense presentation in Section~\ref{sec:projection} would require a
$[\text{batch},\text{sequence},\text{vocabulary}]$ correction tensor. The
production implementation instead observes that every projected direction
remains in the span of the centered expert deltas. Define the per-token Gram
matrix
\begin{equation}
G_{t,mj}
=
\left\langle
\widetilde{\Delta}_{m,t},
\widetilde{\Delta}_{j,t}
\right\rangle.
\label{eq:projection-gram}
\end{equation}
If $d_t=\sum_m a_{m,t}\widetilde{\Delta}_{m,t}$, then
\begin{equation}
\|d_t\|_2^2
=
a_t^\top G_t a_t,
\qquad
\left\langle d_t,\widetilde{\Delta}_{j,t}\right\rangle
=
(a_t^\top G_t)_j.
\label{eq:gram-inner-products}
\end{equation}
A projection against expert $j$ therefore changes only its coefficient,
\begin{equation}
a_{j,t}
\leftarrow
a_{j,t}
-
\operatorname{clip}\!\left(
\frac{(a_t^\top G_t)_j}{G_{t,jj}+\epsilon},
-c_{\max},c_{\max}
\right),
\label{eq:gram-coefficient-update}
\end{equation}
whenever the numerator is negative and the norm guards pass.

Because $d_t$ stays in this span, the protected target remains a linear logit
pool:
\begin{equation}
\log q^{\mathrm{proj}}_t
\overset{+C}{=}
\left(1-\tau\sum_m a_{m,t}\right)\log p_{0,t}
+
\tau\sum_m a_{m,t}\log p_{m,t},
\label{eq:linear-logit-pool}
\end{equation}
where $\overset{+C}{=}$ denotes equality up to an action-independent
constant. The existing OPD kernel can therefore reconstruct the target from
per-token pool coefficients rather than materializing the dense direction.

We compute $G_t$ in two streamed passes over vocabulary blocks: one pass for
the student and expert log-normalizers and one for the raw first and second
moments needed to form the centered Gram matrix. The projection sweep itself
is $O(M^2)$ per token after this computation and stores only the
$M\times M$ Gram matrix, rather than a vocabulary-sized delta. Selective
teacher forwarding is expanded from the routed expert to the union of the
routed and higher-priority constraint experts for each row; lower-priority
experts are never forwarded.

\paragraph{Numerical safeguards.}
Before the sweep, the implementation normalizes $a_t$ by
$\sqrt{a_t^\top G_ta_t}$ and restores its original scale afterward, making
guard behavior invariant to the routed coefficient's magnitude. Reported
runs use centered deltas, $\epsilon=10^{-8}$, an absolute constraint-norm
floor of $10^{-10}$, a relative floor
$G_{t,jj}\geq10^{-6}\|d_t\|_2^2$, $c_{\max}=10$, and a final
direction-preserving rescaling when
$\|\tau a_t\|_1>20$. These guards bound unstable coefficient
representations: the per-step guards skip or cap unsafe corrections, while
the final rescaling preserves the resulting direction.

\section{Experimental Details}
\label{app:experimental-details}

\subsection{Training Protocol}

We study LMOPD on mathematical reasoning using a proprietary internal training
set. All policies start from the same post-SFT checkpoint: a roughly
30B-parameter mixture-of-experts model with approximately 3B active parameters
per token. \ificlrfinal
The checkpoint is an internal predecessor of Cohere's North Mini Code
\citep{cohere_north_code_mini}.
\else
The model architecture is closely related to a subsequently released
open-weight model; additional details will be provided in the de-anonymized
version.
\fi The checkpoint has completed general pretraining and supervised
post-training on multiple domains, including mathematical reasoning, code, and
instruction following. We train and evaluate with a maximum context length of
32K tokens.

Because LMOPD distillation dynamics differ from RLVR and with the number of experts to be distilled from, we run 40 training steps for all 2 expert LMOPD variants, and 80 steps for 4 expert LMOPD variants. Each
step uses 64 prompts with 8 freshly sampled rollouts per prompt, giving 512
fully on-policy trajectories per update. We only filter out empty responses. We use a two-stage learning-rate
schedule: $4\times10^{-6}$ for the first 40 steps and $4\times10^{-7}$
thereafter. All reward-specialist and direct multi-reward RL runs use a common 300-step horizon and their final step-300 checkpoints, with additional filters for zero-variance groups.

\subsection{Evaluation Protocol}

Our final evaluation suite contains AIME 2024, AIME 2025, and HMMT February
2025. AIME is a standard short-answer competition benchmark used extensively
in recent LLM reasoning reports, including Qwen3 \citep{yang2025qwen3}. HMMT
February 2025 provides a distinct set of recent competition problems from the
Harvard--MIT Mathematics Tournament \citep{hmmt2025feb} and is included in the
MathArena evaluation ecosystem \citep{balunovic2025matharena}. None of these
benchmarks is used for training or checkpoint selection. We evaluate each
problem with 50 stochastic rollouts at temperature $0.6$. Repeated rollouts
reduce Monte Carlo noise for each problem but do not create additional
independent benchmark items, so uncertainty estimates and significance tests
treat problems, rather than individual sampled completions, as the primary
evaluation units. We note that public competition benchmarks can be vulnerable
to pretraining contamination; MathArena reports evidence of this issue for
AIME 2024 \citep{balunovic2025matharena}, motivating the inclusion of the more
recent HMMT set and appropriately scoped claims.

\paragraph{Uncertainty and table conventions.}
Retained-gain SEs use the delta method for the ratio, including the shared-base
term. In the four-expert results, the shared base has five replicates, while
the reasoning-correctness specialist endpoint has four because one replicate
lacks AIME-2024/HMMT judge data. We omit an SE for the macro-average because
its axes are dependent. Except for the multi-seed experiment in
\Cref{sec:multiseed-replication}, all reported policies come from one training
seed and their SEs reflect evaluation sampling only.

\clearpage
\subsection{Raw Objective Scores}

\begin{table}[H]
\centering
\caption{Raw objective scores for the evaluated configurations and reference
policies associated with
\Cref{tab:retained_gain_2expert_composite,tab:retained_gain_4expert_composite},
including the intermediate GDPO $(5,1,1)$ tuning runs.
All objectives are composites over AIME 2024, AIME 2025, and HMMT February
2025. Entries are mean $\pm$ SEM over five evaluation-suite replicates. The
RQ-Correctness Expert's RQ subaxes use four replicates. Every Conc column
reports raw $S_{\mathrm{conc}}$. Correctness-Gated RLVR and GDPO reuse the
same policy checkpoints in panels (a) and (b); panel (b) decomposes aggregate reasoning quality into subaxes only for evaluation.}
\label{tab:raw_scores}
\scriptsize
\textbf{(a) Two experts}\par\vspace{2pt}
\resizebox{\textwidth}{!}{%
\begin{tabular}{lccc}
\toprule
\textbf{Model} & \textbf{pass@1} & \textbf{RQ} & \textbf{Conc} \\
\midrule
Rewarded Soup (uniform; 1:1)
& $0.7311 \pm 0.0009$ & $0.7501 \pm 0.0010$ & $0.8038 \pm 0.0002$ \\
Rewarded Soup (correctness-emphasized; 2:1)
& $0.7349 \pm 0.0013$ & $0.7618 \pm 0.0008$ & $0.7973 \pm 0.0003$ \\
Correctness-Gated RLVR (1,1,1)
& $0.7340 \pm 0.0025$ & $0.7392 \pm 0.0032$ & $0.7987 \pm 0.0009$ \\
GDPO (1,1,1)
& $0.5592 \pm 0.0049$ & $0.6397 \pm 0.0016$ & $0.8858 \pm 0.0004$ \\
GDPO (5,1,1)
& $0.6074 \pm 0.0069$ & $0.6734 \pm 0.0028$ & $0.8623 \pm 0.0007$ \\
GDPO (25,1,1)
& $0.7004 \pm 0.0044$ & $0.7174 \pm 0.0010$ & $0.8104 \pm 0.0009$ \\
\rowcolor{LMOPDShade}
\textbf{LMOPD}
& $0.7450 \pm 0.0019$ & $0.7715 \pm 0.0006$ & $0.7980 \pm 0.0003$ \\
\midrule
Base
& $0.7109 \pm 0.0027$ & $0.7186 \pm 0.0029$ & $0.7847 \pm 0.0004$ \\
Reasoning Quality + Correctness Expert
& $0.7440 \pm 0.0043$ & $0.7698 \pm 0.0019$ & $0.7959 \pm 0.0007$ \\
Conciseness Expert
& $0.7111 \pm 0.0044$ & $0.7082 \pm 0.0016$ & $0.8131 \pm 0.0005$ \\
\bottomrule
\end{tabular}%
}

\vspace{6pt}
\textbf{(b) Four experts}\par\vspace{2pt}
\resizebox{\textwidth}{!}{%
\begin{tabular}{lccccc}
\toprule
\textbf{Model} & \textbf{Composite pass@1} & \textbf{RQ-corr} &
\textbf{RQ-verif} & \textbf{RQ-eff} & \textbf{Composite Conc} \\
\midrule
Rewarded Soup (uniform; 1:1:1:1)
& $0.7236 \pm 0.0053$ & $0.7843 \pm 0.0031$ & $0.6469 \pm 0.0016$ & $0.6830 \pm 0.0017$ & $0.8071 \pm 0.0007$ \\
Rewarded Soup (correctness-emphasized; 2:1:1:1)
& $0.7233 \pm 0.0041$ & $0.7910 \pm 0.0052$ & $0.6560 \pm 0.0027$ & $0.6698 \pm 0.0011$ & $0.8005 \pm 0.0002$ \\
Correctness-Gated RLVR (1,1,1)
& $0.7340 \pm 0.0025$ & $0.7866 \pm 0.0033$ & $0.6545 \pm 0.0017$ & $0.6860 \pm 0.0057$ & $0.7987 \pm 0.0009$ \\
GDPO (1,1,1)
& $0.5592 \pm 0.0049$ & $0.6052 \pm 0.0051$ & $0.4237 \pm 0.0005$ & $0.7759 \pm 0.0013$ & $0.8858 \pm 0.0004$ \\
GDPO (5,1,1)
& $0.6074 \pm 0.0069$ & $0.6644 \pm 0.0066$ & $0.5010 \pm 0.0026$ & $0.7540 \pm 0.0029$ & $0.8623 \pm 0.0007$ \\
GDPO (25,1,1)
& $0.7004 \pm 0.0044$ & $0.7583 \pm 0.0026$ & $0.6002 \pm 0.0028$ & $0.6932 \pm 0.0026$ & $0.8104 \pm 0.0009$ \\
\rowcolor{LMOPDShade}
\textbf{LMOPD}
& $0.7476 \pm 0.0042$ & $0.8053 \pm 0.0024$ & $0.6405 \pm 0.0017$ & $0.6396 \pm 0.0021$ & $0.8003 \pm 0.0005$ \\
\midrule
Base
& $0.7109 \pm 0.0027$ & $0.7551 \pm 0.0047$ & $0.6501 \pm 0.0031$ & $0.6429 \pm 0.0032$ & $0.7847 \pm 0.0004$ \\
RQ-Correctness Expert
& $0.7515 \pm 0.0048$ & $0.8117 \pm 0.0035$ & $0.6609 \pm 0.0035$ & $0.6304 \pm 0.0028$ & $0.7919 \pm 0.0003$ \\
RQ-Verification Expert
& $0.7432 \pm 0.0033$ & $0.8255 \pm 0.0030$ & $0.7902 \pm 0.0017$ & $0.5612 \pm 0.0032$ & $0.7602 \pm 0.0009$ \\
RQ-Efficiency Expert
& $0.6285 \pm 0.0049$ & $0.6881 \pm 0.0036$ & $0.5008 \pm 0.0019$ & $0.7827 \pm 0.0026$ & $0.8402 \pm 0.0009$ \\
Conciseness Expert
& $0.7111 \pm 0.0044$ & $0.7694 \pm 0.0043$ & $0.5834 \pm 0.0007$ & $0.6732 \pm 0.0015$ & $0.8131 \pm 0.0005$ \\
\bottomrule
\end{tabular}%
}
\end{table}

\subsection{Multi-Seed Replication}
\label{sec:multiseed-replication}

We replicate full two-expert LMOPD (routing plus priority-aware projection) at
the fixed step-40 endpoint with three independent training seeds. All runs
start from the same SFT checkpoint and use the same configuration; only
\texttt{run.seed} changes, which changes RL-dataloader ordering and therefore
the rollout and routing sequence. Each trained model is evaluated with 50 replicates of the AIME 2024, AIME 2025, and Feb HMMT 2025 benchmarks, identical to other runs. Every
other reported policy uses the original seed, 13207799.

\begin{table}[H]
\centering
\caption{Three-seed replication of full two-expert LMOPD at step 40. Raw
objective entries are mean $\pm$ SEM; RG denotes retained gain. Seed 1 is the run reported in the
main tables and is the median seed by average retained gain. Between-seed SD
measures training-run variation across the three trained models.}
\label{tab:multiseed_2expert}
\scriptsize
\resizebox{\textwidth}{!}{%
\begin{tabular}{lccccccc}
\toprule
\textbf{Training seed} & \textbf{Composite pass@1} & \textbf{Composite RQ} &
\textbf{Composite Conc} & \textbf{pass@1 RG} & \textbf{RQ RG} &
\textbf{Conc RG} & \textbf{Average RG} \\
\midrule
\rowcolor{LMOPDShade}
Seed 1 (13207799; median, main-table run)
& $0.7450 \pm 0.0042$ & $0.7715 \pm 0.0014$ & $0.7980 \pm 0.0006$
& $102.9\%$ & $103.3\%$ & $46.9\%$ & $84.4\%$ \\
Seed 2 (24681357)
& $0.7405 \pm 0.0032$ & $0.7720 \pm 0.0016$ & $0.7989 \pm 0.0002$
& $89.4\%$ & $104.3\%$ & $50.0\%$ & $81.2\%$ \\
Seed 3 (98765431)
& $0.7507 \pm 0.0039$ & $0.7760 \pm 0.0033$ & $0.7997 \pm 0.0007$
& $120.2\%$ & $112.1\%$ & $52.8\%$ & $95.1\%$ \\
\midrule
\textbf{Three-seed mean}
& \textbf{0.7454} & \textbf{0.7732} & \textbf{0.7989}
& \textbf{104.2\%} & \textbf{106.6\%} & \textbf{49.9\%}
& \textbf{86.9\%} \\
Between-seed SD
& 0.0051 & 0.0025 & 0.0009 & 15.4 pt & 4.8 pt & 3.0 pt & -- \\
\bottomrule
\end{tabular}%
}
\end{table}

\begin{figure}[H]
\centering
\includegraphics[width=0.9\textwidth]{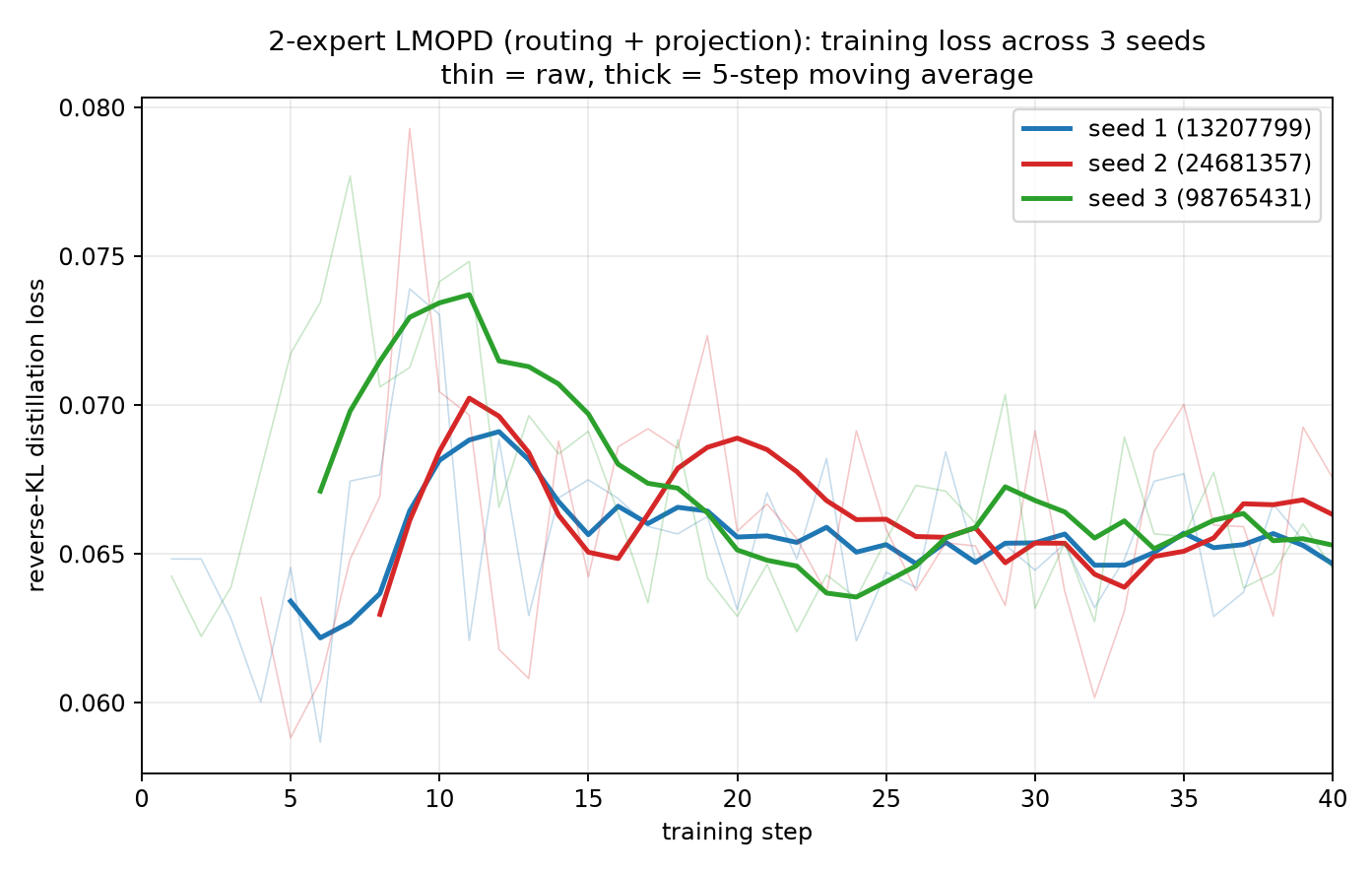}
\caption{Reverse-KL distillation loss for the three two-expert LMOPD training
runs. Thin lines show per-step loss and thick lines show a five-step moving
average. All three runs converge to similar loss values by the fixed step-40
evaluation endpoint.}
\label{fig:multiseed_loss}
\end{figure}

\subsection{Reward Definitions}
\label{app:reward-details}

We use one verifiable outcome reward, four process-quality axes, and a
length-based conciseness reward.

\paragraph{Correctness.}
$R_C(x,y)\in\{0,1\}$ indicates whether the model's final answer exactly matches
the gold answer after answer normalization.

\paragraph{Reasoning quality.}
A frozen 31B Gemma 4 judge model
\citep{gemmateam2026gemma4technicalreport} evaluates four axes:
\emph{reasoning correctness} (whether the derivation steps are valid),
\emph{verification} (whether the response checks or validates its answer),
\emph{clarity} (whether the reasoning is understandable and well organized),
and \emph{efficiency} (whether the reasoning avoids unnecessary dead ends). The full prompt for the judge is included in \cref{app:judge-prompts}
For each axis $a$, the judge emits a penalty
\begin{equation}
    p_a \in \{0,0.3,0.7,1.0\},
    \qquad R_a = 1-p_a,
\end{equation}
where larger reward is better. We define aggregate reasoning quality as
\begin{equation}
    R_{\mathrm{RQ}}
    = \frac{1}{4}
    \left(
    R_{\mathrm{RC}}+R_{\mathrm{RV}}+R_{\mathrm{RCl}}+R_{\mathrm{RE}}
    \right),
\end{equation}
for reasoning correctness, verification, clarity, and efficiency,
respectively.

\paragraph{Conciseness.}
Let $L(y)$ be the number of completion tokens. We first define an intrinsic
conciseness score
\begin{equation}
S_{\mathrm{conc}}(y)
=
1-\lambda\frac{L(y)}{K_L+L(y)},
\label{eq:conc-score}
\end{equation}
with $K_L=16{,}384$ and $\lambda=0.5$. This bounded penalty discourages
unnecessary length without making long but correct solutions arbitrarily
poor. All reported Conc metrics evaluate $S_{\mathrm{conc}}$ directly,
independent of correctness. The conciseness specialist is instead trained with the
correctness-gated reward to mitigate reward hacking:
\begin{equation}
R_{\mathrm{conc}}(x,y)
=
R_C(x,y)\,S_{\mathrm{conc}}(y),
\label{eq:conc-reward}
\end{equation}
so an incorrect short answer cannot obtain a high specialist-training reward.
GDPO is the exception during training: its conciseness dimension is the raw
$S_{\mathrm{conc}}$ score as the objective weights should serve to mitigate reward hacking.

\subsection{Specialist Training and Selection}
\label{app:specialist-details}

We train candidate specialists from the same base checkpoint for 300 RL steps,
separately targeting correctness, each of the four reasoning-quality subaxes,
aggregate reasoning quality, and conciseness. This specialization stage
intentionally isolates objectives before any capability integration is
attempted. We retain specialists based on their target-reward validation
performance rather than assuming one usable expert per reward. In pilot
specialist training, the specialists for aggregate reasoning quality and
reasoning correctness are stronger capability teachers than the binary-correctness
specialist, consistent with the denser feedback supplied by the process
reward. We therefore allow these experts to serve the correctness gate as
well. We omit a separate clarity expert from the integration experiments
because specialist training does not provide a meaningful benchmark
improvement on clarity over the starting model.

\subsection{Baseline Objectives}
\label{app:baseline-details}

\paragraph{GDPO.}
For each reward dimension $i$, GDPO independently computes a group-normalized
advantage
\begin{equation}
    A_i = \frac{r_i-\mu_i}{\sigma_i+\epsilon},
\end{equation}
aggregates the reward-specific advantages,
\begin{equation}
    \widetilde A = \sum_{i=1}^{K} w_i A_i,
\end{equation}
and applies a final batch-wise normalization before the policy-gradient
update. Unlike LMOPD, this combines all reward signals into a single
policy-gradient direction. In our implementation, the conciseness reward
dimension is raw $S_{\mathrm{conc}}$, without a correctness gate; we evaluate
equal weights and $5{:}1$ and $25{:}1$ emphases on correctness. GDPO receives
the same three dimensions---$R_C$, aggregate $R_{\mathrm{RQ}}$, and raw
$S_{\mathrm{conc}}$---in both experimental settings, so each weighting reuses
the same trained checkpoint; the four-expert table decomposes
$R_{\mathrm{RQ}}$ only for evaluation.

\paragraph{Correctness-Gated RLVR.}
The correctness-gated objective used in both settings is
\begin{equation}
R_{\mathrm{multi}}(x,y)
=
\frac{
R_C(x,y)\left(
    w_c + w_r R_{\mathrm{RQ}}(x,y)
    + w_k S_{\mathrm{conc}}(y)
\right)
}{w_c+w_r+w_k}.
\label{eq:gated-multitask}
\end{equation}
The gate prevents an incorrect response from being compensated by aggregate
reasoning quality or conciseness scores, but lower-priority objectives are
still scalarized together once correctness is satisfied. Because
$R_{\mathrm{RQ}}$ already averages reasoning correctness, verification,
clarity, and efficiency, the same trained policy is evaluated in both
experimental settings; the four-expert results report those subaxes
separately only for diagnosis.

\section{Qualitative Case Study}
\label{app:qualitative}

\begin{figure}[H]
\centering
\includegraphics[width=\textwidth]{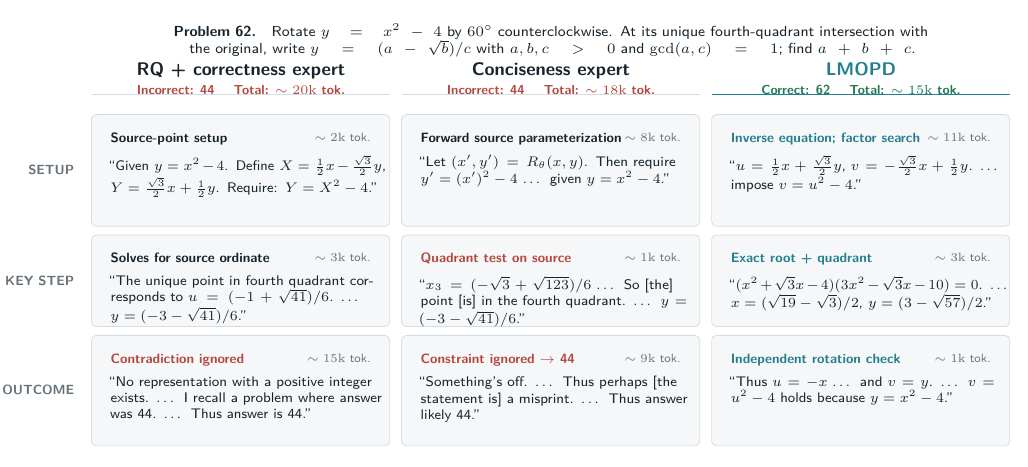}
\caption{\textbf{Contrastive trajectories on Problem 62.} We deliberately
select completed trajectories that expose distinct run-level outcomes: the Reasoning Quality + Correctness Expert performs an extended search and
encounters evidence that its candidate answer is inconsistent, but does not
revise it, leading to the wrong answer. The Conciseness Expert reaches an answer more directly, but an early
setup error remains unchecked, leading to the same wrong answer. LMOPD abandons an unproductive intermediate
path, recovers the correct answer, and validates it through an independent
check. In the
original ten-rollout trace sets, exact-answer counts are 4/10, 6/10, and 8/10,
respectively. Ellipses
and bracketed text denote omissions and light editing, respectively.}
\label{fig:problem-62-qualitative}
\end{figure}

\section{LLM Judge Prompts}
\label{app:judge-prompts}

We use the following prompt to evaluate the reasoning quality of each
generated response. The full generated response is sent afterwards as a user message.

\begin{llmprompt}[Reasoning Quality Judge Prompt]
You are judging the REASONING-PROCESS QUALITY of a math solution.

You do NOT know the official gold answer. Do NOT score whether the final number matches gold.
Score HOW the solution reasons: whether it derives a coherent finished answer, and how verified,
clear, and efficient that reasoning is.

The candidate solution may contain two parts:
- [Reasoning]: internal scratch work / exploration
- [Final answer]: the answer shown to the user
Evaluate BOTH. A flaw anywhere counts, including in [Reasoning], even if [Final answer] looks clean.

Score exactly FOUR criteria. Assign each its OWN penalty in {0.0, 0.3, 0.7, 1.0}:
  0.0 = no flaw on this criterion
  0.3 = minor flaw
  0.7 = major flaw
  1.0 = fatal flaw
Judge criteria independently: a failure on one must not automatically raise the others, and a
clean final answer must not automatically clear the style criteria.

Split of labor (keep these boundaries sharp):
- reasoning_correctness = derivation + answer validity (sound, justified, complete, supported,
  consistent, not guessed). This is NOT gold-correctness.
- verification_self_check / conceptual_clarity / efficiency_conciseness = STYLE among solutions
  that already look derived. Prefer verified, early-clear, high-signal paths over unchecked,
  opaque, or bloated ones.
Boundary examples:
- Answer obtained by guessing / recall / vibes with no valid derivation -> reasoning_correctness
  (not efficiency).
- Valid derivation, but via an unnecessarily long grind / abandoned detours -> efficiency
  (and possibly conceptual_clarity), not reasoning_correctness alone.
- Post-answer self-check (or its absence) -> verification_self_check only; never "buy" a lower
  reasoning_correctness penalty with a check, and never treat a check as fixing an inconsistent
  or unsupported answer.

CRITERIA

1. reasoning_correctness
   ONE penalty for whether the solution DERIVES a coherent, finished answer. Cover ALL facets
   below; set the penalty to the severity of the WORST material flaw (do not average a fatal
   flaw down with milder ones):
   (a) Soundness - algebraic / arithmetic / symbolic / geometric / probabilistic / theorem steps
       correct? A wrong intermediate that propagates IS a flaw.
   (b) Justification - major claims derived, or unexplained jumps / miracle formulas?
   (c) Completeness - necessary cases, domains, boundaries, degeneracies, extraneous roots handled?
   (d) Answer support / completion - clear final answer (e.g. well-formed \\boxed{...} when
       required) actually supported by the preceding reasoning? Incomplete / non-terminating /
       stuck traces: at least 0.7. No real answer: 1.0.
   (e) Internal consistency - [Reasoning] and [Final answer] agree? Contradictions, or talking
       oneself into a different number after failed attempts without a repaired derivation: 1.0.
       Mid-trace self-corrections that ARE repaired into a consistent derivation do not by
       themselves force a high penalty here.
   (f) Rigor vs guessing - answer DERIVED, or guessed / recalled / vibes ("I recall", "plausibly",
       guess-and-check used in place of a proof)? Luck with no valid derivation: 0.7-1.0.
   Defaults: a clean, fully derived, internally consistent writeup is 0.0 here even if long,
   opaque, or unverified (those costs belong on criteria 2-4). A confident-looking but
   inconsistent, unsupported, or guessed answer is usually 0.7-1.0 here.

2. verification_self_check
   AFTER a candidate answer is derived, did the solution run an INDEPENDENT check that could have
   FALSIFIED a wrong answer? Be STRICT.
   Award 0.0 ONLY for a clearly separate post-answer check that uses a method DIFFERENT from the
   derivation path (second method; reverse derivation from the claimed answer; bound / parity /
   mod check not already implied by the derivation steps).
   Same-path rechecking is NOT independent (at least 0.3), including: plugging candidates back
   into the same equation/condition that produced them; re-doing the same arithmetic; re-deriving
   with the same method "as a check"; enumerating the same finite cases already counted;
   numerically approximating a sum already obtained in closed form.
   Also insufficient for 0.0 (at least 0.3 even if present): mid-trace self-corrections without a
   final independent check; "so the answer is N" / confidence language; a special case that does
   not stress the claimed general answer.
   Defaults: typical LM AIME writeup -> 0.3 unless the independent-check bar is clearly met; no
   check at all -> 0.7. Do not award 0.0 merely because the number looks confident.

3. conceptual_clarity
   Is the KEY INSIGHT named as a strategy early and then used, or buried in opaque algebra / case
   soup? Be STRICT.
   Award 0.0 ONLY if, near the start (before most of the calculation), the solution states a short
   strategy/principle a reader could quote in 1-2 sentences, then executes that plan.
   At least 0.3 when: the idea is clear only after most of the grind; the idea is never named
   (only executed); insight appears after long exploration; or the "insight" merely restates the
   setup. 0.7 when the path is an opaque case/coordinate grind with no recoverable principle.
   Naming a principle only after long trial does NOT earn 0.0.
   Defaults: typical correct LM writeup -> 0.3 unless the early-named-strategy bar is clearly met.
   Do not confuse with verbosity: a short opaque grind is worse than a slightly longer explanation
   that leads with the idea.

4. efficiency_conciseness
   Is the trace an efficient, high-signal path to the answer? Penalize:
   (a) Bloat / repetition - repeated calculations or phrases; re-deriving the same fact;
       redundant answer restatement / boxing padding: at least 0.3. Doom loops or near-duplicated
       blocks that dominate the trace: 1.0.
   (b) Path economy - brute force, exhaustive casework, or exploratory grind when a short
       structural argument (symmetry, invariant, closed form, clever substitution) would do.
       (If there is no valid derivation at all, that is reasoning_correctness (f), not only this.)
   (c) Claim density - throat-clearing, problem restatement, or non-advancing paraphrase; large
       deletable stretches. Abandoned approaches more than about one-third of the reasoning: at
       least 0.3; majority of the trace: at least 0.7.
   Full re-derivation of the same result "as a check" costs efficiency here and does NOT earn
   verification_self_check = 0.0.
   Defaults: crisp, mostly load-bearing path -> 0.0; mild padding or mild over-casework -> 0.3;
   heavy bloat / brute force / low signal -> 0.7-1.0. Do not penalize length if every part does
   necessary work - penalize wasted, repeated, or structurally inefficient work.

CALIBRATION
- Use only the anchors {0.0, 0.3, 0.7, 1.0}. If you can point to a concrete flaw, give a nonzero
  penalty (do not be conservative).
- If uncertain between two adjacent penalties on ANY criterion, choose the HIGHER one.
- Standard correct, complete LM AIME writeup (default prior):
    reasoning_correctness = 0.0
    each of verification_self_check, conceptual_clarity, efficiency_conciseness = 0.3
  unless that style criterion clearly meets its 0.0 bar. Expect nonzero on at least TWO of the
  three style criteria. Reserve 0.0 on ALL three style criteria only for rare textbook traces:
  independent different-method check + early named strategy + high load-bearing fraction.
- Before clearing all three style criteria to 0.0, re-check: (i) is the "second method" truly
  different, or just finishing/enumerating the same argument? (ii) is there answer-format padding,
  repeated summary, or exploratory aside? If unclear, keep at least 0.3 on the weak style criterion.
- Incomplete / non-terminating / doom-loop traces: reasoning_correctness usually 0.7-1.0, and also
  penalize verification_self_check and efficiency_conciseness.
- Confident tone does not excuse unsupported formulas, inconsistent answers, or guessing.

Return JSON ONLY, with a penalty AND a one-clause reason for every criterion:
{
  "criteria": {
    "reasoning_correctness":         {"penalty": 0.0, "reason": ""},
    "verification_self_check":       {"penalty": 0.0, "reason": ""},
    "conceptual_clarity":            {"penalty": 0.0, "reason": ""},
    "efficiency_conciseness":        {"penalty": 0.0, "reason": ""}
  },
  "worst_criterion": "reasoning_correctness | verification_self_check | conceptual_clarity | efficiency_conciseness | none",
  "brief_reason": "One sentence naming the dominant reasoning weakness, or 'Clean, direct, verified solution.'"
}
\end{llmprompt}

\end{document}